\documentclass{article}

\usepackage[main, final]{neurips_2026}

\usepackage[utf8]{inputenc} 
\usepackage[T1]{fontenc}    
\usepackage{hyperref}       
\hypersetup{hidelinks}
\usepackage{url}            
\usepackage{booktabs}       
\usepackage{amsfonts}       
\usepackage{nicefrac}       
\usepackage{microtype}      
\usepackage{xcolor}         
\usepackage{amsmath}
\usepackage{graphicx}

\newsavebox{\crossmodeltablebox}
\newcommand{\rwpara}[1]{\vspace{0.35em}\noindent\textbf{#1.}\ }
\title{Beyond Solving: Prescriptive Probing for Neural Routing Solvers}

\author{%
  Reuben Narad\thanks{Corresponding author: \texttt{rnarad@uw.edu}} \\
  Foster School of Business\\
  University of Washington\\
  Seattle, WA, USA \\
  \texttt{rnarad@uw.edu} \\
  \And
  L\'eonard Boussioux \\
  Foster School of Business\\
  Paul G.\ Allen School of Computer Science \& Engineering\\
  University of Washington\\
  Seattle, WA, USA \\
  \And
  Michael Wagner \\
  Foster School of Business\\
  University of Washington\\
  Seattle, WA, USA
}

\begin{document}

\maketitle
\enlargethispage{2\baselineskip}




\begin{abstract}
Neural combinatorial optimization (NCO) trains fast heuristics for routing problems, but planners often need more than a single solve: they ask which stop to drop, which transition to preserve, or which subset of stops to remove if a route is infeasible. Answering such counterfactual questions by re-solving each candidate is expensive even when the action set is small. We introduce \textit{prescriptive probing}: using the frozen representations of a trained NCO model to rank candidate interventions for what-if decision support. While recent interpretability work descriptively probes what NCO solvers encode, we ask whether those same representations can prescriptively guide intervention choices on real-road routing problems. Where exhaustive labels are available, we train supervised probes from offline re-solve labels; where they are not (e.g., combinatorial action spaces), we train a lightweight reinforcement learning intervention head. Across real-road benchmarks on the Asymmetric Traveling Salesperson Problem and Capacitated Vehicle Routing Problem, probes built on frozen model representations achieve the strongest performance on a majority of intervention tasks and improve over local-repair heuristics on edge-forbiddance and on key node- and multi-node-removal metrics. We further investigate probeability as a property of the routing model itself, varying model quality, the mix of supervised learning and reinforcement learning, and the architecture family. Overall, our results suggest a new use case for NCO solvers: representations learned for route construction can transfer to decision support.
\end{abstract}

\section{Introduction}
\label{sec:introduction}

Routing problems such as the Traveling Salesperson Problem (TSP), its asymmetric variant (ATSP), and the Capacitated Vehicle Routing Problem (CVRP) are central to logistics, and neural combinatorial optimization (NCO) has emerged as a promising way to learn fast heuristics for them \citep{bello2016neural,kool2019attention,
kwon2020pomo,kim2022symnco,kwon2021matnet,luo2023lehd,sun2023difusco}. NCO models are typically evaluated as direct solvers: given an instance, the model constructs or improves a route, and is judged by route cost. Many real planning workflows, however, also involve questions about a route once it is in hand: which stop to drop, which planned transition is most critical to preserve, or which subset of stops to remove when a route becomes infeasible? Each such question is a counterfactual, and answering it by brute force means modifying the instance, re-solving under each candidate intervention, and comparing the resulting costs, which is expensive even when the action set is only linear in the number of stops.

We ask whether trained NCO models can be useful beyond solving. Specifically, can their frozen internal representations serve as operational signals for downstream what-if decision support? This perspective is inspired by pragmatic uses of model internals in other domains: activation probes can flag harmful or policy-relevant behavior in production large language models \citep{gdm2026production_probes}, and frozen Evo 2 genomic foundation-model embeddings can support lightweight variant-effect and pathogenicity predictors in a label-scarce scientific domain
\citep{goodfire2025evee}. In routing, the analogous question is whether a model trained to solve routing instances has learned representations that make costly counterfactual decisions easier to rank.

We call this idea \emph{prescriptive probing}. Recent NCO probing work uses frozen solver representations descriptively, asking what learned solvers encode and how those encodings explain routing behavior \citep{zhang2025probingnco}. Prescriptive probing, in contrast, asks whether those representations can guide actions. Given a trained solver, we freeze its representations and train a small head to rank downstream interventions by their counterfactual value under the target solver. The resulting probe provides a cheap screening mechanism: score many candidate interventions in a single model forward pass, then optionally re-solve only a shortlist.

\begin{figure}[t!]
    \centering
    \makebox[\linewidth][c]{%
        \includegraphics[width=1\linewidth]{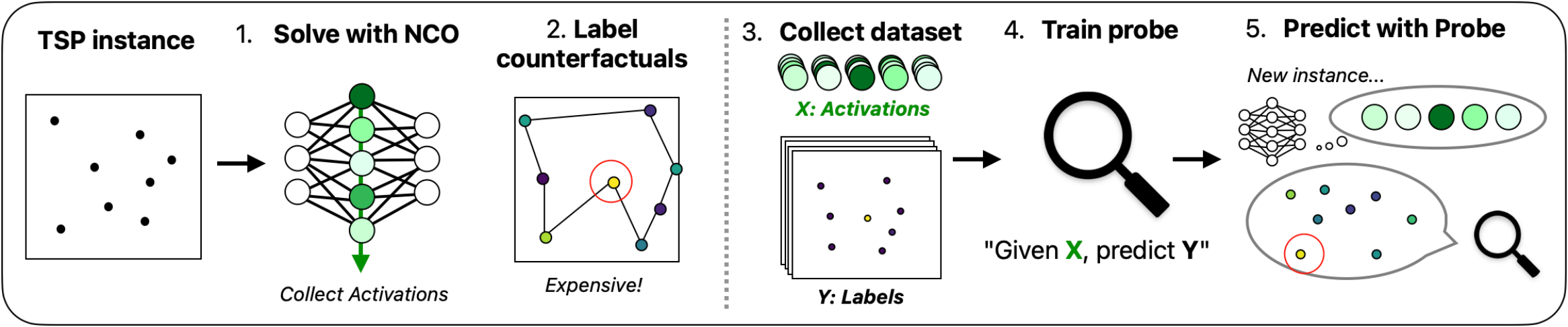}%
    }
    \caption{
    Overview of prescriptive probing. A trained NCO model first solves a routing instance and exposes frozen internal activations. Offline, we label counterfactual interventions by modifying the instance, re-solving with a target solver, and recording the resulting action values. A lightweight probe is then trained to map activations to these labels. At test time, the probe scores candidate interventions for a new instance using a single forward pass through a routing model; the top-ranked actions can be selected directly or used as a shortlist for a small number of expensive re-solves.
    }
    \label{fig:flow}
\end{figure}

We study this framework in real-road asymmetric routing. Unlike synthetic Euclidean TSP, real road networks induce explicit asymmetric origin-destination cost matrices because of one-way streets, turn restrictions, topological constraints, and direction-dependent travel times. We therefore build on Real Routing NCO (RRNCO) and its real-road city benchmark: RRNCO is designed for asymmetric distance and duration matrices, and its row/column encoder produces node-aligned incoming and outgoing representations before the decoder commits to a route
\citep{son2025rrnco}. Euclidean TSP remains useful as a controlled sandbox for comparing alternative model families, but the headline setting of this paper is real-road routing, where operational what-if decisions are most natural.

Our experiments consider three intervention regimes. In \emph{single-node removal}, the action is to remove one stop and re-solve the reduced instance.
In \emph{single-edge forbiddance}, the action is to forbid one directed transition from a planned route and measure the cost of the best
alternative. Both tasks have enumerable candidate sets, so we collect offline re-solve labels and train supervised prescriptive probes. In \emph{multi-node removal}, the action is a subset of stops; exhaustive labels are infeasible because the action space is combinatorial, so we train a lightweight
reinforcement-learning intervention head on the same frozen representations. Across tasks, we compare against simple operations research (OR)-style local repair heuristics and
learned controls that do not use trained model representations.

Empirically, methods using frozen routing representations improve learned controls on the main ATSP intervention benchmarks and outperform local repair heuristics on key node-remove, edge-forbid, and multi-node-removal metrics. We also find that probeability is shaped by the routing model itself: prescriptive signal varies across encoder layers, training checkpoints, and supervised/reinforcement learning (SL/RL) training mixtures. Finally, a Euclidean TSP100 architecture sandbox shows that prescriptive probeability is not unique to RRNCO and varies substantially across modern NCO representation families. Overall, our results suggest that NCO models can be useful not only when they produce the final route, but also when their learned states provide reusable operational signals for expensive downstream decisions.

\section{Background and Related Work}
\label{sec:background}

\rwpara{Neural combinatorial optimization and routing}
Neural combinatorial optimization trains neural models as fast heuristics for routing problems such as TSP, ATSP, CVRP, and related variants. Early work framed routing as sequence prediction or RL \citep{vinyals2015pointer,bello2016neural}, while the Attention Model became a standard Transformer-style encoder-decoder baseline for learned routing \citep{kool2019attention}. Subsequent autoregressive and constructive methods improved training, symmetry handling, matrix encodings, and decoder-side computation \citep{kwon2020pomo,kim2022symnco,kwon2021matnet,luo2023lehd}.
Other paradigms expose different representation objects: GCN and heatmap methods predict edge scores, local-search methods learn guidance signals, and diffusion or test-time optimization methods operate through edge- or denoising-centric states \citep{joshi2019efficient,fu2021generalize,joshi2022gnnlocal,
sun2023difusco,li2023t2t,li2024fastt2t}. A related line uses learning as a component of an explicit solver, for example, by guiding destroy-repair, large-neighborhood search, ant-colony search, or subproblem selection \citep{chen2019learning,hottung2020neural_lns,li2021delegate,ye2024glop,
deepaco2023}. Recent generalist and unified approaches broaden NCO across problems and distributions \citep{drakulic2025goal,pan2025unico, li2025unifyml4tsp,ma2025ml4cobench}. These works primarily ask how well neural methods solve routing problems; we ask whether the representations learned while solving can also support downstream what-if decisions.

\rwpara{Real-road asymmetric routing}
Most learned routing benchmarks use synthetic Euclidean TSP or CVRP, where the coordinates induce symmetric distances. Earlier matrix-encoding work showed that neural CO models can operate directly on relationship matrices, including ATSP \citep{kwon2021matnet}. RRNCO extends this direction to real-world routing, using asymmetric distance and duration matrices and exposing row/column representations for incoming and outgoing structures \citep{son2025rrnco}. This makes RRNCO a natural source of representation for real-road ATSP what-if tasks. We use Euclidean TSP100 only as a controlled-architecture sandbox for model families that do not natively support explicit asymmetric real-road matrices.

\rwpara{Probing and pragmatic representation reuse}
Probing trains lightweight predictors on frozen representations to test what information is accessible from model states \citep{alain2016linear,adi2016fine,conneau2018probing,belinkov2022probing}.
This connects to mechanistic interpretability, which studies internal computation in neural networks, and to a pragmatic view in which model internals are useful when they support concrete downstream tasks such as monitoring, classification, steering, or scientific prediction
\citep{elhage2021transformer_circuits,anthropic2023monosemanticity, nanda2025pragmatic,gdm2026production_probes,goodfire2025evee}. Recent NCO probing work applies this idea descriptively, asking whether learned solvers encode distances, demands, route membership, myopia-avoidance information, constraint structure, or layer-wise decision-relevant features \citep{zhang2025probingnco}. Our use is prescriptive: rather than asking only what a routing model encodes, we ask whether those frozen representations can rank operational interventions.

\rwpara{What-if optimization, reoptimization, and interdiction}
OR has long studied routing problems where solutions change under local
instance modifications. TSP reoptimization considers node insertion/deletion
\citep{archetti2003reoptimizing,ausiello2009reoptimization}; interdiction and fortification study edge disruptions or increased arc costs
\citep{lozano2017interdiction}; and orienteering, prize-collecting, and quota variants model optional service over subsets of customers
\citep{vansteenwegen2011orienteering,bienstock1993prize}. Our intervention labels follow this tradition: modify the instance or planned route, re-solve with a target solver, and measure the resulting objective. The bottleneck is label cost: even \(O(n)\) action sets require many re-solves, while subset interventions are combinatorial.

\section{Methodology: Prescriptive Tasks and Probing}
\label{sec:methodology}

\subsection{Counterfactual interventions and solver-relative labels}
\label{sec:method_counterfactuals}

Let \(x=(V,C)\) be a routing instance with stops \(V=\{1,\ldots,n\}\) and, in
the ATSP setting, an asymmetric cost matrix \(C\in\mathbb{R}^{n\times n}\). Let
\(\mathcal{S}\) be the target solver used to evaluate interventions. The
framework is solver-agnostic: \(\mathcal{S}\) may be an exact solver, heuristic,
commercial optimizer, or learned routing model, depending on what an
organization uses in deployment.

A prescriptive task specifies candidate actions \(\mathcal{A}(x)\). Each action
\(a\in\mathcal{A}(x)\) induces a modified instance \(x^a\), whose value is
obtained by re-solving with \(\mathcal{S}\). We write \(J_{\mathcal{S}}(x)\) for
the route cost returned by \(\mathcal{S}\), \(q_{\mathcal{S}}(a;x)=
J_{\mathcal{S}}(x^a)\) for the counterfactual value, and
\(\Delta_{\mathcal{S}}(a;x)=J_{\mathcal{S}}(x^a)-J_{\mathcal{S}}(x)\) for the
effect relative to the base solution. Labels are therefore \emph{solver-relative}: an intervention is important relative to the
counterfactual behavior of the chosen solver. Even when \(|\mathcal{A}(x)|\) is
only \(O(n)\), fully labeling one instance requires many counterfactual re-solves.

\subsection{Frozen routing representations}
\label{sec:method_representations}

Let \(M_\phi\) be a trained routing model. We reuse internal states of
\(M_\phi\) as frozen representations for what-if prediction, writing
\(H_\phi^p(x)\) for the representation extracted at encoder probe point \(p\).
For RRNCO, which is an encoder-decoder model for real-road routing, the encoder
builds node-aligned representations from coordinates and real-road edge
features, including asymmetric distance and duration information
\citep{son2025rrnco}. The decoder then uses the encoded instance and partial
route to construct a solution autoregressively.

We probe encoder residual-stream states because they are computed once per
instance and remain aligned with input nodes before any particular route is
decoded. This follows recent NCO probing work, which also reads node embeddings
across AM/POMO encoder layers and LEHD decoder states
\citep{zhang2025probingnco}. At probe point \(p\), RRNCO exposes row states
\(\{r_i^p\}_{i\in V}\) and column states \(\{c_i^p\}_{i\in V}\). Our default
node representation is \(h_i^p=[r_i^p;c_i^p]\), where row and column streams
capture outgoing- and incoming-oriented structure. We sweep over early,
intermediate, and final encoder probe points; unless otherwise stated, \(M_\phi\)
is frozen and only the probe or intervention head is trained.

\begin{figure*}[t]
    \centering
    \makebox[\linewidth][c]{%
        \includegraphics[width=1\linewidth]{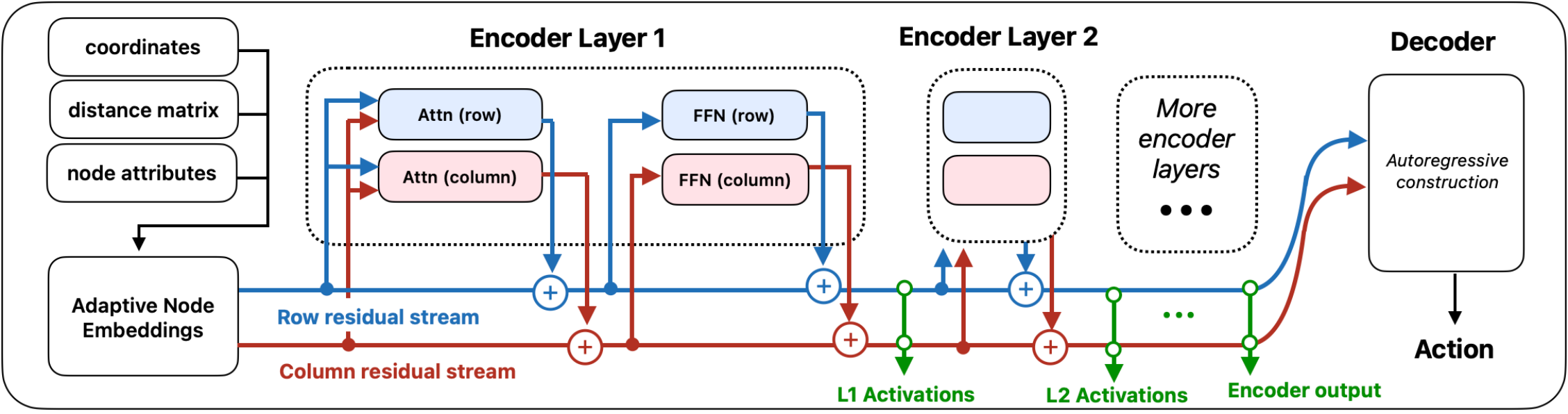}%
    }
    \caption{
    Representation pipeline for prescriptive probing with RRNCO. The routing
    model encodes a real-road instance using separate row and column residual
    streams, which remain aligned with the input stops. We extract frozen states
    from multiple encoder probe points, such as early, intermediate, and final
    row/column residual-stream locations. Node probes score candidate removals
    from row/column node states; edge probes score used directed transitions
    from source and target endpoint states; and remove-\(k\) intervention heads
    sequentially select subsets using the same frozen representations. The
    routing model is not updated during probe or intervention-head training.
    }
    \label{fig:arch}
\end{figure*}

\subsection{Supervised single-intervention tasks}
\label{sec:method_supervised_interventions}

We first study two \(O(n)\)-candidate intervention tasks (single-node removal and single-edge forbiddance) that can be labeled
offline by exhaustive counterfactual re-solving. For both, a probe receives
frozen representations from \(M_\phi\) at probe point \(p\) and predicts one
score per candidate action, written as
\(\hat{q}_\theta(a;x,p)=g_\theta(H_\phi^p(x),a)\). Probes rank candidates by predicted impact on tour length. Probe-head details are in Appendix~\ref{app:probe_heads}.

\paragraph{Single-node removal.}
Each candidate action removes one node:
\(\mathcal{A}_{\mathrm{node}}(x)=V\), or
\(V\setminus\{\mathrm{depot}\}\) in depot-based settings. Let \(x^{-i}\) be the
instance with node \(i\) removed. The best node is
\(i_{\mathcal{S}}^\star(x)=\arg\min_{i\in\mathcal{A}_{\mathrm{node}}(x)}
J_{\mathcal{S}}(x^{-i})\). This models decisions such as last-mile triage: a
driver has 100 planned deliveries but, because of space or time constraints, can
serve only 99, so the planner must choose one package to drop. It is also a
natural reoptimization problem: TSP reoptimization under node addition/deletion
is NP-hard \citep{archetti2003reoptimizing,ausiello2009reoptimization}.

\paragraph{Single-edge forbiddance.}
This is a post-solve task: given a base route \(\tau_{\mathcal{S}}(x)\), we ask
which used directed transition is most critical to preserve. For TSP or ATSP,
the candidates are successor pairs
\(\mathcal{A}_{\mathrm{edge}}(x,\tau_{\mathcal{S}})
=\{(\tau_t,\tau_{t+1}):t=1,\ldots,n\}\). For a used edge \((u,v)\), let
\(x^{-(u,v)}\) forbid \(v\) from immediately following \(u\). The target is
\((u,v)_{\mathcal{S}}^\star(x)=
\arg\max_{(u,v)\in\mathcal{A}_{\mathrm{edge}}(x,\tau_{\mathcal{S}})}
[J_{\mathcal{S}}(x^{-(u,v)})-J_{\mathcal{S}}(x)]\). This models route-level
contingency analysis: if time-sensitive operations depend on certain planned
transitions, which ones should be monitored most closely for disruption? While a real road closure usually changes many origin–destination costs rather than forbidding a single successor pair, the edge-forbid task provides a controlled abstraction: it tests whether the representations that support node-removal sensitivity also capture localized pairwise route sensitivity.

\begin{figure}[t]
    \centering
    \makebox[.9\textwidth][c]{%
        \begin{tabular}{@{}c@{}}
            \includegraphics[width=1\linewidth]{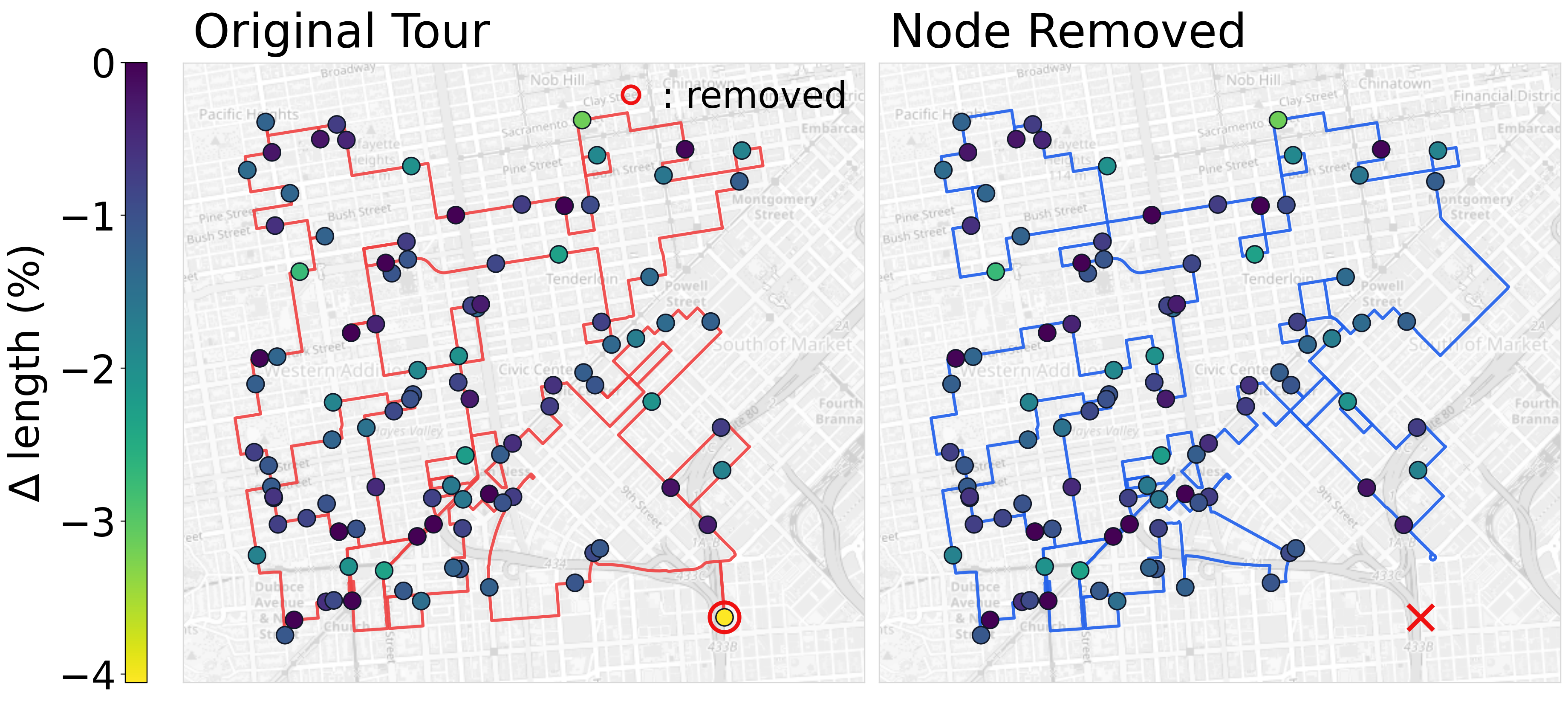}\\
            \hspace{.13cm}\includegraphics[width=.98\linewidth]{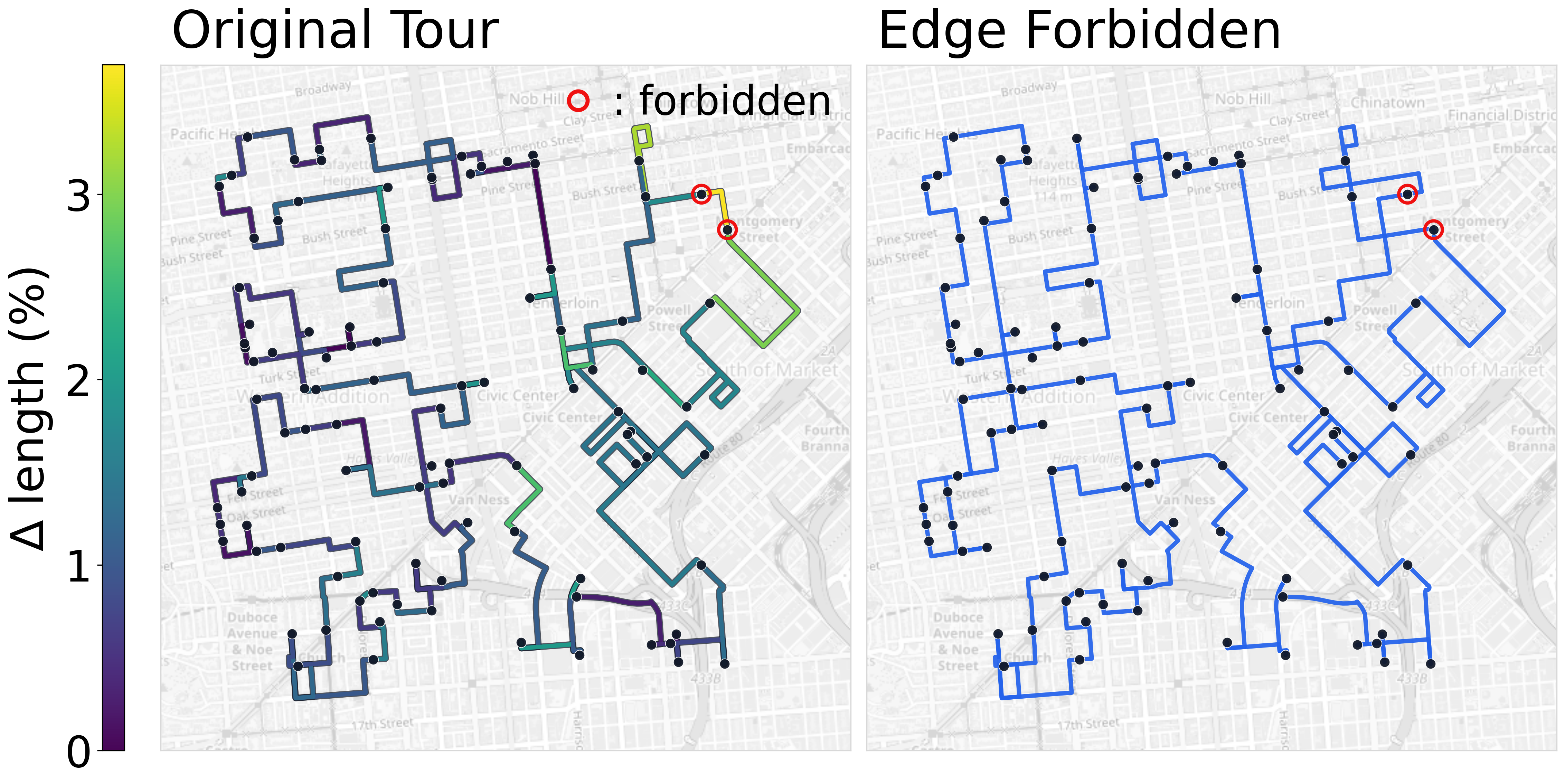}
        \end{tabular}%
    }
    \caption{
    Two ATSP instances from San Francisco illustrating the single-node
    removal (top) and edge-forbid (bottom) tasks. Candidate actions are colored by the change in route cost induced by
    removing/forbidding and re-solving; the highlighted node/edge is the most impactful. Additional examples of both tasks are
    provided in Appendix~\ref{app:additional_task_examples}.
    }
    \label{fig:task_examples}
\end{figure}

\subsection{Multi-node removal}
\label{sec:method_multi_node}

We also study a combinatorial intervention, remove-k: choose \(k\) stops to remove,
\(\mathcal{A}_k(x)=\{R\subseteq V: |R|=k\}\). This models cases where a route is
substantially infeasible and several stops must be dropped. The ideal target is
\(R_{\mathcal{S},k}^\star(x)=\arg\min_{R\subseteq V,\ |R|=k}
J_{\mathcal{S}}(x^{-R})\), but exhaustive labels require \(\binom{n}{k}\)
re-solves. This illustrates a broader limitation of supervised prescriptive
probing: many useful interventions have clear counterfactual rewards but no
tractable exhaustive label set.

In routing, however, the reward can still be queried by trying a proposed
intervention and re-solving. We therefore train a lightweight online
intervention head rather than a supervised probe. The head sequentially selects
nodes without replacement,
\(i_t\sim \pi_\theta(\cdot \mid H_\phi^p(x),i_1,\ldots,i_{t-1})\), and receives
reward after \(x^{-\{i_1,\ldots,i_k\}}\) is re-solved. This use of RL is natural
in OR-style domains, where candidate actions can be simulated or optimized
against a black-box solver, even when exhaustive labels are unavailable. The
task tests whether frozen routing representations support combinatorial
intervention policies, not only independent single-action scores.

\subsection{Baselines and metrics}
\label{sec:method_baselines_metrics}

We compare against two baseline families. \emph{Heuristic baselines} approximate
counterfactual effects by cheap local route repair: for node removal, delete a
node and shortcut its predecessor to its successor; for edge forbiddance,
replace a forbidden transition with a local detour. These are simple but strong. \emph{Learned controls} use the same probe-head families as
the representation probes, but replace trained activations with raw routing
inputs such as coordinates, cost matrices, and summary features. These controls
test whether performance comes from trained routing states rather than probe
capacity or raw instance information alone. Appendix~\ref{app:heuristic_results}
additionally reports random-action references.

For supervised tasks, the main benchmark reports Acc@1, Acc@5, Reg@1, and
Reg@5. Acc@1 is the fraction of instances where the probe's top-ranked action
equals the true best action; Acc@5 is the fraction where the true best action
appears anywhere in the probe's top five. Reg@1 measures the regret of the
top-ranked action, while Reg@5 measures the regret after taking the best action among the probe's top five candidates. Thus Acc@5 and
Reg@5 correspond to a screening workflow: the probe shortlists five
interventions and the planner re-solves only that shortlist. We additionally use
Spearman rank correlation in diagnostic analyses where full-ranking agreement is
useful. For remove-\(k\), we report the resulting tour reduction after selected
nodes are removed and the instance is re-solved. Metric definitions are in
Appendix~\ref{app:metric_details}, and implementation details are in
Appendix~\ref{app:experiment_details}.


\section{Experimental Setup}
\label{sec:experiments}

We organize the experiments around three questions. First, how well do
prescriptive probes perform on real-road intervention tasks, compared with
learned controls and simple local-repair heuristics? Second, what aspects of a
routing model make it probeable, across training checkpoints and
NCO architectures? Third, can the probing analysis give practical
guidance for training routing models, for example in choosing the amount of
supervised pretraining before RL continuation?

\subsection{Main real-road benchmark}
\label{sec:experiments_main}

The main benchmark uses real-road instances of 80 cities from the RRNCO data pipeline. We evaluate ATSP100 and ATSP500, with 100 and 500 stops, and include RCVRP100 customer removal as a capacitated-routing extension. ATSP instances have no depot; RCVRP instances include a depot, customer demands, vehicle capacity, and customer-removal candidates only.

We evaluate three intervention regimes. Node removal resolves the instance after removing a stop. Edge forbiddance resolves after forbidding one used directed edge in a planned ATSP route; we omit RCVRP edge forbiddance because depot/customer route edges require a separate task definition. Remove-5, the k = 5 instance of remove-k, trains a policy to sequentially remove five ATSP nodes and receives a reward after the reduced instance is re-solved.

All real-road counterfactual labels are defined with respect to Lin–Kernighan-style (LKH)
heuristic solvers \citep{helsgaun2000lkh,helsgaun2017lkh3}. This matches the
solver-relative framing of Section~\ref{sec:method_counterfactuals}: the probe
learns intervention value under the target solver used to generate labels.
Node-removal labels are \(100(L_{\mathrm{base}}-L_{\mathrm{removed}})/
L_{\mathrm{base}}\), edge-forbid labels are
\(100(L_{\mathrm{forbid}}-L_{\mathrm{base}})/L_{\mathrm{base}}\), and remove-5
uses the same tour-improvement reward after removing the selected nodes.
Dataset sizes, splits, solver trials, and timeouts are in
Appendix~\ref{app:experiment_details}.

\subsection{Routing models, probes, and controls}
\label{sec:experiments_models}

The main representation source is an RRNCO checkpoint trained on the
corresponding real-road problem family. For ATSP100 and ATSP500, we use
real-road ATSP RRNCO models initialized with weak-LKH supervised training and
continued with RL; RCVRP100 uses the corresponding capacitated RRNCO model. We
extract row/column encoder states at two probe locations, an intermediate
layer-2 state and the final encoder state, as described in
Section~\ref{sec:method_representations}. Full architecture and routing-model
training details are given in Appendix~\ref{app:experiment_details}.

For each supervised problem-task pair, we run a validation sweep over
probe location, learning rate, input type, and probe head. Input types are
\emph{raw}, using the same coordinates, cost information, and summary features
available to the routing model; \emph{activation}, using frozen RRNCO states;
and \emph{raw+activation}, concatenating both. Probe heads include linear, standard MLP, and RowCol MLP variants, where
RowCol MLP keeps the RRNCO row and column streams separate through the first
projection before scoring their interaction; it is a probe-head choice rather
than an additional raw cost-matrix input.
All supervised probes use a ranking-aware soft
cross-entropy loss and are selected by validation Regret@1. The main table
reports the selected probe for each setting, with final results averaged over 10 seeds. For remove-5, we train analogous lightweight RL intervention heads that
select nodes sequentially without replacement and select by validation reward.
Full swept configurations, feature definitions, and optimization details are in
Appendix~\ref{app:experiment_details} and Appendix~\ref{app:probe_heads}.

\subsection{Probeability analyses}
\label{sec:experiments_probeability}

Beyond the fixed-model benchmark, we ask what makes a routing model more
probeable. We run two analyses on RRNCO training. First, we track probe
performance over RL training for an ATSP100 large model, extracting checkpoints
throughout RL continuation and training both linear and row-column MLP probes
for node removal and edge forbiddance. This tests whether prescriptive signal
improves as the same routing model becomes a better solver.

Second, we run an iso-performance and supervised-learning/reinforcement-learning (SL/RL) sweep: starting from a fully supervised
RRNCO model, we resume earlier supervised checkpoints with RL until matched route
performance, then train linear and MLP node-removal probes. This gives models
with comparable routing quality but different SL/RL mixtures. A small amount of
SL before RL improves probeability across both probe families, with the clearest
effect for linear probes and a consistent Spearman trend for MLP probes.

Finally, we evaluate released NCO checkpoints in Euclidean TSP100, enabling a
controlled comparison across model families that do not all support explicit
asymmetric real-road matrices. We include autoregressive, symmetry-augmented,
heatmap, heavy-decoder, diffusion, and RRNCO-style models, probing each model's
most natural internal state. We compare node-removal probe accuracy against
TSP100 policy gap; provenance and probe-location details are in
Appendix~\ref{app:cross_arch_details}.

\section{Results and Analysis}
\label{sec:results}

\subsection{Main real-road results}
\label{sec:results_main}

Table~\ref{tab:main_results} reports the main real-road benchmark. ATSP100
node removal is the cleanest activation-only result: the layer-2 RowCol MLP
reaches 69.8\% Acc@5 and reduces Reg@5 from 1.13 for the raw-input control to
0.23. The local repair heuristic remains strongest on exact top-1 and Reg@1,
showing that learned representations are most useful here as a high-quality
screening signal rather than a universal replacement for local tour repair.

The larger and constrained node-removal settings are more baseline-competitive.
On ATSP500, raw inputs remain strongest on accuracy and Reg@1, while
raw+activation slightly improves Reg@5; on RCVRP100, local repair remains very
strong on regret, while raw+activation gives the best Acc@5 among learned rows.
Edge forbiddance is the clearest win over local repair: raw+activation probes
improve both accuracy and regret over heuristic and raw-input baselines on
ATSP100 and ATSP500. Remove-5 shows the same mixed picture: raw+activation RL
wins on ATSP100, while local top-5 deletion remains strongest on ATSP500.
Overall, frozen routing states provide useful prescriptive information, but the
best use of that signal depends on task, scale, and the strength of local
repair.

\begin{table*}[t]
\centering
\scriptsize
\setlength{\tabcolsep}{3.0pt}
\renewcommand{\arraystretch}{0.94}
\caption{
Main real-road benchmark results. Acc@5/1 reports top-5 and top-1 accuracy in percentage points;
Reg@5/1 reports regret at top-5 and top-1. Remove-5 reports the average tour
improvement. Configurations were chosen by a sweep over probe location, learning rate, and probe type;
Local-repair baselines are deterministic. Means are over 10
seeds; See Appendix~\ref{app:full_main_results} for details.
}
\label{tab:main_results}
\resizebox{\linewidth}{!}{%
\begin{tabular}{l l l l l c c c}
\toprule
Problem & Task & Input & Location & Probe Type
& Acc@5 / Acc@1 \(\uparrow\)
& Reg@5 / Reg@1 \(\downarrow\)
& Tour Reduction (\%) \(\uparrow\) \\
\midrule
ATSP100 & node-remove & heuristic & N/A & local repair
        & 65.5 / \textbf{42.4} & 0.41 / \textbf{1.12} & -- \\
        &             & raw & N/A & MLP
        & 30.9 / 11.7 & 1.13 / 2.76 & -- \\
        &             & activation & L2 & RowCol MLP
        & \textbf{69.8} / 36.1 & 0.23 / 1.36 & -- \\
        &             & raw+activation & L2 & MLP
        & 69.6 / 34.9 & \textbf{0.23} / 1.38 & -- \\
\midrule
ATSP500 & node-remove & heuristic & N/A & local repair
        & 19.5 / 9.1 & 1.18 / 2.88 & -- \\
        &             & raw & N/A & MLP
        & \textbf{34.1 / 17.2} & 0.92 / \textbf{2.63} & -- \\
        &             & activation & Final & RowCol MLP
        & 25.0 / 12.9 & 1.08 / 2.90 & -- \\
        &             & raw+activation & Final & MLP
        & 32.9 / 16.7 & \textbf{0.92} / 2.70 & -- \\
\midrule
RCVRP100 & node-remove & heuristic & N/A & local repair
         & 7.3 / \textbf{2.2} & \textbf{6.05 / 11.53} & -- \\
         &             & raw & N/A & MLP
         & 8.0 / 1.8 & 6.06 / 11.88 & -- \\
         &             & activation & Final & RowCol MLP
         & 7.1 / 2.0 & 6.23 / 12.09 & -- \\
         &             & raw+activation & L2 & MLP
         & \textbf{8.1} / 2.1 & 6.06 / 11.74 & -- \\
\midrule
ATSP100 & edge-forbid & heuristic & N/A & local repair
        & 20.7 / 9.0 & 2.71 / 4.80 & -- \\
        &             & raw & N/A & MLP
        & 27.9 / 12.8 & 2.52 / 4.25 & -- \\
        &             & activation & L2 & RowCol MLP
        & 25.8 / 12.3 & 2.55 / 4.24 & -- \\
        &             & raw+activation & L2 & MLP
        & \textbf{29.7 / 15.9} & \textbf{2.37 / 3.87} & -- \\
\midrule
ATSP500 & edge-forbid & heuristic & N/A & local repair
        & 20.9 / 10.6 & 1.43 / 2.32 & -- \\
        &             & raw & N/A & MLP
        & 41.1 / 17.7 & 1.11 / 1.95 & -- \\
        &             & activation & L2 & RowCol MLP
        & 37.8 / 16.3 & 1.16 / 1.98 & -- \\
        &             & raw+activation & Final & MLP
        & \textbf{46.0 / 21.5} & \textbf{1.03 / 1.72} & -- \\
\midrule
ATSP100 & remove-5 & heuristic & N/A & local top-5
        & -- & -- & 12.3 \\
        &          & raw & N/A & MLP policy
        & -- & -- & 12.0 \\
        &          & activation & Final & RowCol policy
        & -- & -- & 13.7 \\
        &          & raw+activation & Final & MLP policy
        & -- & -- & \textbf{14.7} \\
\midrule
ATSP500 & remove-5 & heuristic & N/A & local top-5
        & -- & -- & \textbf{7.8} \\
        &          & raw & N/A & MLP policy
        & -- & -- & 7.2 \\
        &          & activation & Final & RowCol policy
        & -- & -- & 6.4 \\
        &          & raw+activation & L2 & MLP policy
        & -- & -- & 7.3 \\
\bottomrule
\end{tabular}%
}
\end{table*}

\vspace{-.1cm}
\subsection{Probeability tracks routing-model quality}
\label{sec:results_probeability}

\begin{figure*}[t]
    \centering
    \makebox[\linewidth][c]{%
        \includegraphics[width=0.9\linewidth]{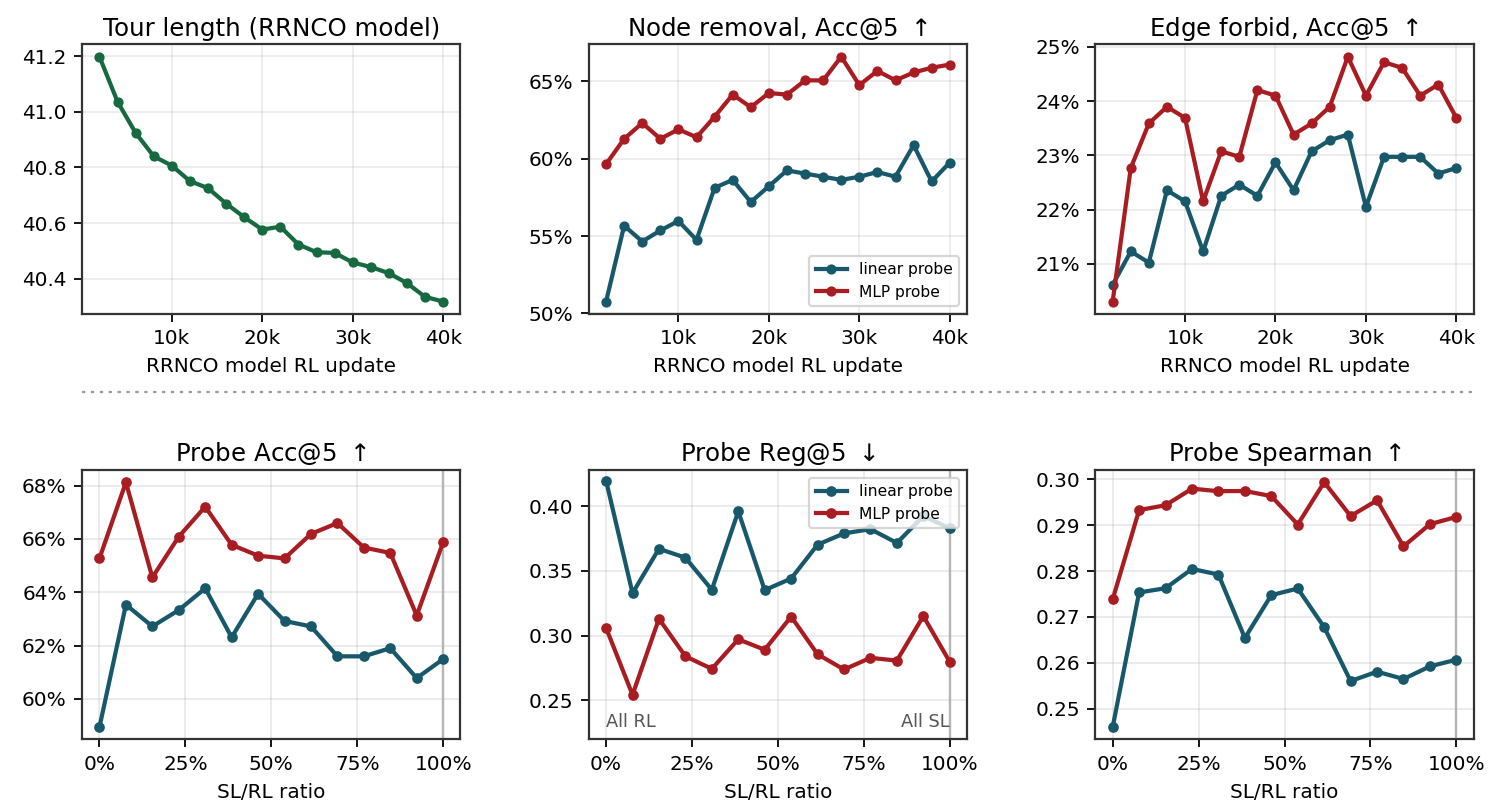}%
    }
    \caption{
    Probeability analyses. Top: probe performance during RL continuation of the
    ATSP100 RRNCO model. As the routing model improves, downstream
     probe accuracy also improves for both probe types. Bottom:
    iso-performance SL/RL sweep for node-removal probes. Fixing routing quality, a small amount of SL before RL yields the best
    probeability; the effect is strongest for linear probes.
    }
    \label{fig:probeability_figs}
\end{figure*}

We show that probeability is shaped by the routing model itself.
Figure~\ref{fig:probeability_figs} (top) tracks the ATSP100 RRNCO foundation
model used in the main benchmark during the RL training stage. As performance
improves, both node-removal and edge-forbid Acc@5 improve. The row-column MLP remains stronger, but both probe
families follow the same upward trend. Thus, within a fixed architecture and
training trajectory, better routing models expose more useful prescriptive
signal.

\subsection{A little SL goes a long way}
\label{sec:results_isoperf}

Figure~\ref{fig:probeability_figs} (bottom) shows the iso-performance SL/RL
sweep: we compare checkpoints with matched route quality but different amounts
of supervised pretraining before RL continuation. Neither the pure-RL nor the
pure-SL endpoint is baest. Instead, a small amount of supervised pretraining,
roughly 10\%--30\% of the supervised anchor, followed by mostly RL yields the strongest probeability. The effect is visible in both linear and MLP probes.

\vspace{-.1cm}
\subsection{Cross-architecture analysis}
\label{sec:results_cross_model}

\begin{figure}[t]
    \centering
    \sbox{\crossmodeltablebox}{%
        \scriptsize
        \setlength{\tabcolsep}{3.0pt}
        \renewcommand{\arraystretch}{1.4}
        \begin{tabular}{@{}l l c c c@{}}
            \toprule
            Model & Family & Gap & Acc@1 & Acc@5 \\
            \midrule
            DIFUSCO {\scriptsize(Sun and Yang\ 2023)}
                & Diffusion
                & 0.24 & \textbf{50.0} & \textbf{93.3} \\[4pt]
            Sym-NCO {\scriptsize(Kim et al.\ 2022)}
                & \shortstack[l]{AR constructive}
                & 0.38 & 36.7 & 83.3 \\[4pt]
            RRNCO-ATSP {\scriptsize(Son et al.\ 2026)}
                & \shortstack[l]{AR constructive}
                & 3.82 & \textbf{50.0} & 83.3 \\[4pt]
            GCN-Heatmap {\scriptsize(Joshi et al.\ 2019)}
                & Non-AR
                & 1.57 & 43.3 & 73.3 \\[4pt]
            POMO {\scriptsize(Kwon et al.\ 2020)}
                & \shortstack[l]{AR constructive}
                & 0.37 & 33.3 & 70.0 \\[4pt]
            AM {\scriptsize(Kool et al.\ 2019)}
                & \shortstack[l]{AR constructive}
                & 4.57 & 20.0 & 60.0 \\[4pt]
            LEHD {\scriptsize(Luo et al.\ 2023)}
                & \shortstack[l]{AR constructive}
                & 0.56 & 16.7 & 40.0 \\
            \bottomrule
        \end{tabular}
    }
    \makebox[\textwidth][c]{%
        \begin{minipage}[t]{0.52\textwidth}
            \vspace{0pt}
            \centering
            \usebox{\crossmodeltablebox}
        \end{minipage}
        \hspace{0.02\textwidth}
        \begin{minipage}[t]{0.5\textwidth}
            \vspace{0pt}
            \centering
            \includegraphics[
                height=\dimexpr1.\dimexpr\ht\crossmodeltablebox+\dp\crossmodeltablebox\relax\relax,
                keepaspectratio
            ]{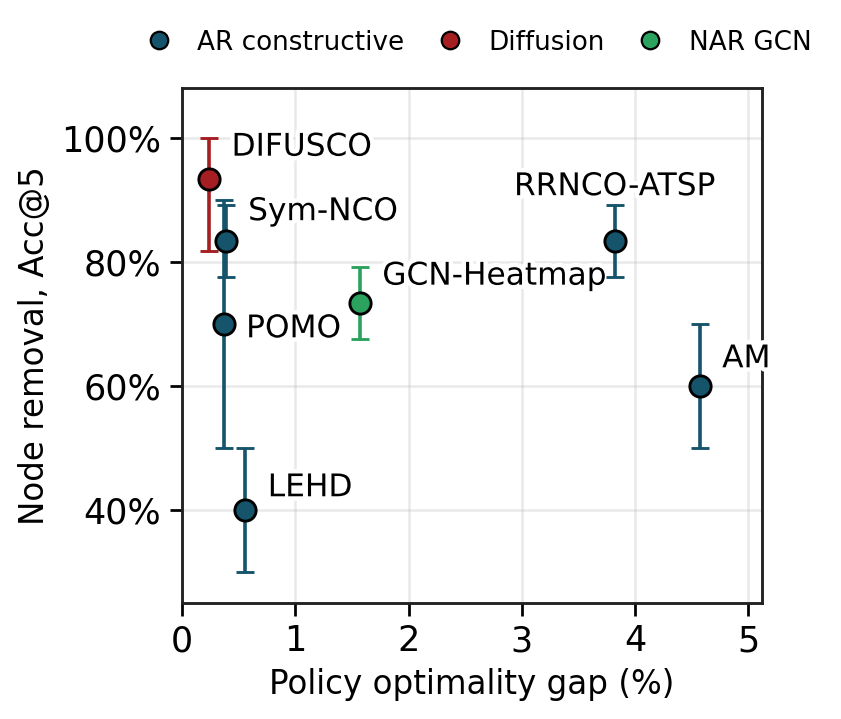}
        \end{minipage}
    }
    \caption{
    Euclidean TSP100 cross-architecture comparison. Left: released-checkpoint
    policy optimality gap and node-removal probe accuracy, averaged over three
    split/probe seeds. Right: policy gap versus Acc@5. We find that diffusion and heatmap-style representations are also
    probeable in Euclidean TSP100. RRNCO-ATSP reference is
    trained for real-road, asymmetric routing.
    }
    \label{fig:cross_model}
\end{figure}

Finally, Figure~\ref{fig:cross_model} presents the Euclidean TSP100
cross-architecture sandbox. This experiment is not the real-road headline: most
modern alternatives to RRNCO are built for 2D Euclidean TSP and do not natively
support explicit asymmetric real-road matrices. The sandbox instead asks whether
prescriptive probeability is unique to RRNCO, and how it varies across
representation families in a domain where released checkpoints are directly
comparable.

The comparison shows that prescriptive probeability extends beyond RRNCO while varying substantially across representation families. DIFUSCO is strongest, pairing a small policy gap with 93.3\% Acc@5; Sym-NCO, GCN-Heatmap, and POMO are also strong. Solver quality alone does not explain the pattern: LEHD is less probeable despite competitive routing performance (consistent with its definitionally light encoder), while RRNCO-ATSP probes well despite being evaluated out of domain. Thus, better solvers often expose better prescriptive signal, but the representation interface also matters. Together, the real-road benchmark and probeability analyses show that prescriptive signal is not merely a property of the downstream probe head. It is shaped by the trained routing representation, the training recipe, and the architecture family exposing it. This supports the central view of the paper: NCO models can be evaluated not only by the routes they produce, but also by the operational usefulness of the internal states they learn while solving.

\vspace{-.1cm}
\section{Discussion, Limitations, and Conclusion}
\vspace{-.1cm}

\label{sec:discussion}

Our results support a view of neural routing models as operationally useful
encoders. Frozen routing representations provide useful signal for real-road
what-if tasks, especially when combined with directly available route and cost
features. The probeability analyses further show that this signal is shaped by
the routing model itself: it improves during RL training, changes under
different SL/RL mixtures at matched route quality across probe families, and varies across NCO architecture families.
Thus, route cost alone does not fully summarize what a
neural routing model has learned; its internal states can also be evaluated by
their usefulness for downstream prescriptive decisions.

The study has several limitations. Labels are solver-relative, so a probe
trained on one routing engine learns that engine's notion of intervention
value; this is appropriate for deployment but limits solver-independent claims.
The labels also remain expensive: prescriptive probing amortizes offline
re-solving, but still requires counterfactual data or simulator-style reward
queries. Our intervention families cover node removal, edge forbiddance, and
multi-node removal, while richer logistics settings include time windows,
stochastic travel times, and route-level disruptions.
Finally, cross-architecture probing is inherently imperfect because different
NCO families expose different internals. Even with these limits, the
results suggest a broader evaluation axis for NCO: not only the route a model
produces, but how useful its learned state is for the operational decisions
surrounding that route.

\bibliographystyle{plainnat}
\bibliography{bibliography}

\appendix

\section{Probe Head Architectures}
\label{app:probe_heads}

This appendix describes the probe heads used for the supervised
single-intervention tasks in Section~\ref{sec:method_supervised_interventions}.
In all cases, the routing model \(M_\phi\) is frozen. Only the probe parameters
are trained.

\paragraph{Node heads.}
For single-node removal, each candidate action corresponds to one node \(i\).
At encoder probe point \(p\), RRNCO provides row and column states \(r_i^p\) and
\(c_i^p\). The simplest node representation is the row-column concatenation
\(h_i^p=[r_i^p;c_i^p]\). A linear probe scores this representation with
\(s_i=w^\top h_i^p+b\), testing whether intervention value is linearly exposed
in the chosen model state.

We also use a row-column interaction MLP. This head first maps row and column
states separately, \(u_i=\phi_r(r_i^p)\) and \(v_i=\phi_c(c_i^p)\), then scores
the interaction vector \(z_i=[u_i;v_i;u_i-v_i;u_i\odot v_i]\), where \(\odot\)
denotes elementwise multiplication. This preserves RRNCO's directional
row/column structure while allowing the probe to use alignment, mismatch, and
multiplicative interactions between incoming- and outgoing-oriented states.

This interaction-augmented input is related in spirit to interaction features
used in prior NCO probing work, including CS-Probing-style analyses
\citep{zhang2025probingnco}, but differs in purpose and architecture:
CS-Probing analyzes statistically significant coefficients of linear probes,
whereas our row-column head is a nonlinear predictor of counterfactual
intervention value.

\paragraph{Edge heads.}
For single-edge forbiddance, each candidate action is a directed used edge
\((u,v)\in\mathcal{A}_{\mathrm{edge}}(x,\tau_{\mathcal{S}})\). The probe input
is built from source and target endpoint representations. A generic edge input
uses \(h_{uv}^p=[h_u^p;h_v^p]\). For RRNCO, a more directionally structured
input is \(h_{uv}^p=[r_u^p;c_v^p]\), since the transition \(u\to v\) combines
the outgoing role of \(u\) with the incoming role of \(v\).

We also evaluate interaction-augmented edge inputs such as
\([h_u^p;h_v^p;|h_u^p-h_v^p|;h_u^p\odot h_v^p]\), and the corresponding
row-column variants. The edge probe outputs a scalar disruption score
\(\hat{\Delta}_{uv}=g_\theta(h_{uv}^p)\). Candidate edges are ranked by this
score, with larger values indicating more critical planned transitions.

\section{Additional task examples}
\label{app:additional_task_examples}

\begin{figure}[t]
    \centering
    \includegraphics[width=\linewidth]{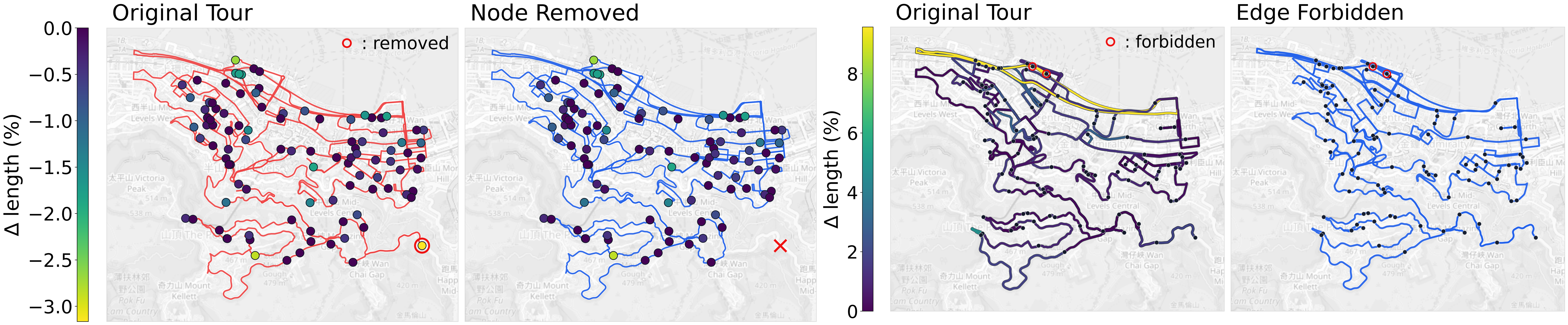}

    \vspace{0.5em}

    \includegraphics[width=\linewidth]{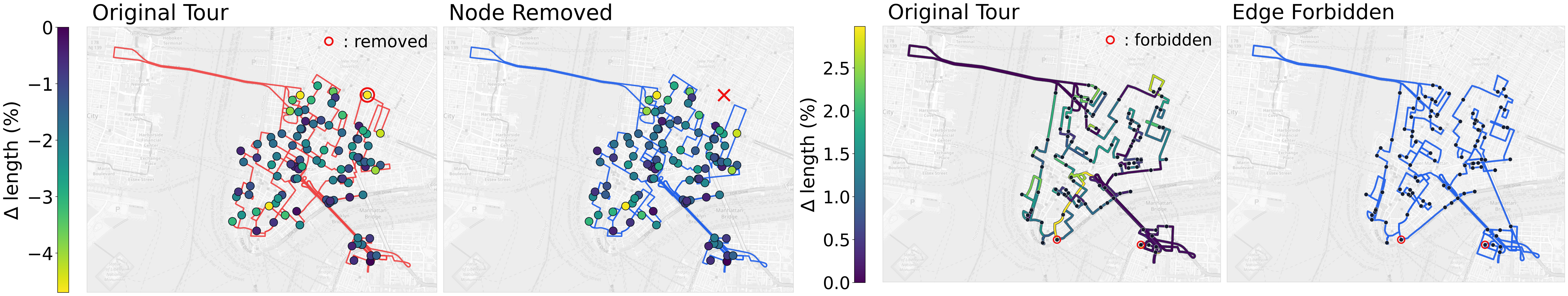}

    \vspace{0.5em}

    \includegraphics[width=\linewidth]{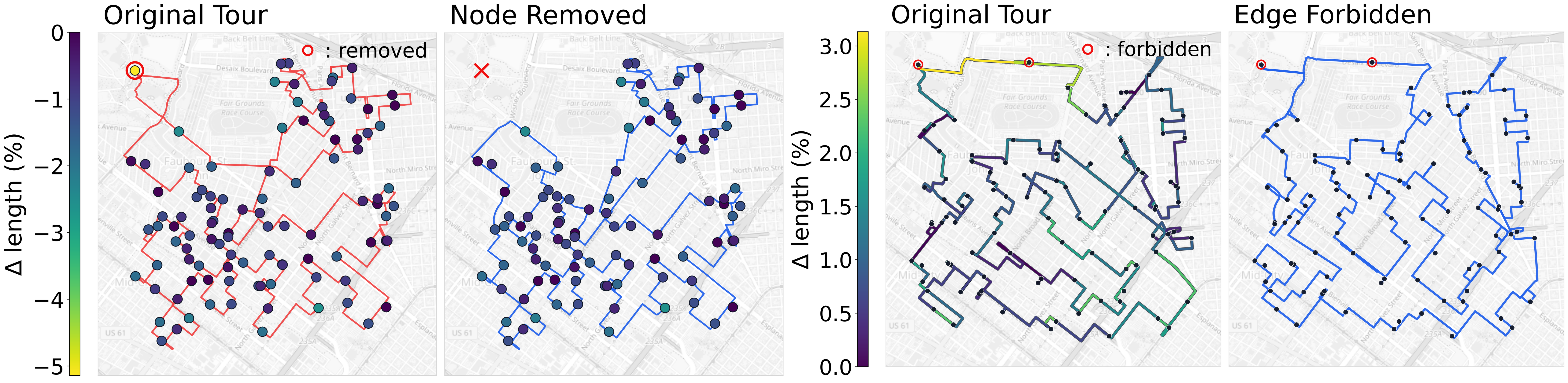}

    \caption{
    Additional real-road task examples for single-node removal and single-edge
    forbiddance. Rows show Hong Kong, New York City, and New Orleans instances.
    In each row, the left two panels show the node-removal task before and after
    removing the highlighted stop, and the right two panels show the edge-forbid
    task before and after forbidding the highlighted directed transition.
    Candidate colors indicate the counterfactual route-cost change obtained by
    applying the intervention and re-solving.
    }
    \label{fig:additional_task_examples}
\end{figure}

Figure~\ref{fig:additional_task_examples} shows additional real-road examples
of the two single-intervention tasks. Each row shows one city instance, with
single-node removal on the left pair of panels and single-edge forbiddance on
the right pair of panels. For node removal, candidate stops are colored by the
change in route cost induced by removing that stop and re-solving; the
highlighted stop is the best removal action. For edge forbiddance, used directed
transitions are colored by the increase in route cost induced by forbidding that
transition and re-solving; the highlighted transition is the most damaging
forbidden edge. The post-intervention panels show the corresponding re-solved
routes after applying the highlighted intervention.

\section{Evaluation Metrics}
\label{app:metric_details}

For supervised intervention tasks, let \(a^\star\) be the best action under the
counterfactual labels and let \(\hat{a}_1,\hat{a}_2,\ldots\) be the predicted
ranking. Acc@1 is the fraction of instances for which
\(\hat{a}_1=a^\star\). Acc@5 is the fraction of instances for which
\(a^\star\in\{\hat{a}_1,\ldots,\hat{a}_5\}\). Spearman is the rank correlation
between predicted scores and true counterfactual action values within each
instance, averaged across instances.

For minimization-style tasks such as node removal, Reg@1 is
\(q_{\mathcal{S}}(\hat{a}_1;x)-q_{\mathcal{S}}(a^\star;x)\), and Reg@5 is the
best regret among the top five predicted actions. For disruption-style tasks
such as edge forbiddance, regret is the gap between the true maximum disruption
and the disruption achieved by the selected candidate. Remove-5 reports average
tour reduction after removing five selected nodes and re-solving:
\(100(L_{\mathrm{base}}-L_{\mathrm{removed}})/L_{\mathrm{base}}\).

\section{Experimental Details}
\label{app:experiment_details}

\paragraph{Real-road datasets and splits.}
The main real-road benchmark uses ATSP100, ATSP500, and RCVRP100 instances from
the RRNCO data pipeline. ATSP100 and ATSP500 contain directed routing instances
with 100 and 500 stops and asymmetric real-road distance matrices. RCVRP100
contains 100 customers plus one depot, customer demands, vehicle capacity, and a
real-road distance matrix. The supervised probe datasets use instance-level
splits. ATSP100 node removal uses 9,783 labeled instances split into
7,826/978/979 train/validation/test examples; ATSP100 edge forbiddance uses
9,757 instances split into 7,806/976/975. Note that the ATSP100 dataset is not 10,000 due to some nodes being unreachable during instance generation, making it unsolvable by LKH. ATSP500 node removal and edge
forbiddance each use 10,000 instances split into 8,000/1,000/1,000. RCVRP100
customer removal uses 10,000 instances split into 8,000/1,000/1,000.

\paragraph{Counterfactual labels.}
For ATSP node removal, each candidate node is removed, the reduced instance is
re-solved with LKH, and the label is
\(100(L_{\mathrm{base}}-L_{\mathrm{removed}})/L_{\mathrm{base}}\). ATSP100
node-removal labels use one LKH run per candidate; ATSP500 node-removal labels
use one run per candidate with a 20-second timeout. RCVRP100 customer-removal
labels use LKH ADCVRP tooling with 10 trials per candidate and a 60-second
timeout. For edge forbiddance, the base LKH route is fixed, each used directed
transition is forbidden, and the constrained instance is re-solved. ATSP100
edge-forbid labels use one LKH run per candidate; ATSP500 edge-forbid labels
use 10 LKH trials per candidate. For remove-5, labels are not enumerated
offline. The policy selects five nodes, the reduced instance is solved online
with LKH, and the reward is the percentage tour improvement after removal.

\paragraph{Representation extraction.}
For ATSP100 and ATSP500, we use 6-layer RRNCO base models with 128-dimensional
row and column streams, giving 256-dimensional concatenated node
representations. For RCVRP100, we use an 8-layer RRNCO model with
192-dimensional row and column streams, giving 384-dimensional concatenated
customer representations. We extract encoder states at the layer-2 and final
probe points. The routing models are frozen for all probe training.

\paragraph{Compute resources.}
All reported training and probing jobs were run on H200 nodes. The RRNCO
foundation runs used approximately 6 hours of supervised learning followed
by 18 hours of RL continuation per run. Remove-5 RL intervention-head runs used
a 30-minute wall-clock budget per run. In the iso-performance SL/RL sweep, RL
continuation from the earlier supervised checkpoints took at most 6 hours to reach parity with the full SL checkpoint. Supervised probe runs
used the fixed epoch and early-stopping budgets described below; counterfactual
label-generation cost is governed by the LKH trials and timeouts described
above.

\paragraph{Probe training and selection.}
The supervised probe datasets are split by instance, typically with about 8k
training instances and 1k each for validation and test. Final main-table probes
use a fixed budget of 240 epochs with validation early stopping patience 40.
The optimizer is AdamW with weight decay \(10^{-4}\), a cosine learning-rate
schedule, and batch size 32. MLP-style heads use hidden dimension 1024, four
layers, and dropout 0.1. Probe configurations are selected by validation
Regret@1 for the main supervised tables, and final results are reported over 10
seeds. Learning rate is included in the validation sweep rather than fixed
analytically.

\paragraph{Raw-input features.}
The raw-input MLP baseline receives problem-specific non-representation
features. For ATSP node removal, these include coordinates, nearest outgoing
and incoming distances, and row/column distance summary statistics. For ATSP
edge forbiddance, they include source and target coordinates, coordinate
differences, directed edge costs, edge position in the tour, and nearest-neighbor
distance summaries. For RCVRP node removal, they include customer coordinates,
demand, depot distances, nearest row/column distances, and row/column summary
statistics. All raw features are standardized using the training split.

\paragraph{Training-dynamics probes.}
The training-dynamics sweeps use lighter probe-training settings: max epoch
budget 80, early-stopping patience 12, learning rate \(3\times 10^{-4}\), batch
size 32, and a cosine learning-rate schedule. The split is the same 80/10/10
instance split as in the main supervised experiments. These probes often stop
well before the maximum epoch budget.

\paragraph{Euclidean cross-architecture probes.}
The Euclidean cross-architecture comparison uses 100 exact-labeled TSP100
instances split 80/10/10. Because the train split contains only 80 instances,
the probe has 8k node-candidate labels. These probes use max epoch budget 120,
early-stopping patience 20, learning rate \(10^{-3}\), batch size 16, and no
learning-rate scheduler. Models are selected by validation Acc@1.

\paragraph{Local repair diagnostics.}
For node removal, the local shortcut heuristic removes node \(i\) and connects
its predecessor directly to its successor in the base route. For edge
forbiddance, the local replacement heuristic estimates the cost of replacing a
forbidden directed transition with a cheap alternative transition or local
detour. These heuristics are used as OR-style diagnostic baselines and as
context for the learned methods; exhaustive counterfactual re-solving is the
label oracle, not a deployable baseline.

\section{Full Main-Benchmark Results}
\label{app:full_main_results}

The main text reports a compact validation-selected subset in
Table~\ref{tab:main_results}. This appendix reports the full supervised and RL
benchmark tables, including all swept learned rows and mean \(\pm\) standard
deviation over 10 seeds. Deterministic local-repair and random-reference
baselines are reported in Appendix~\ref{app:heuristic_results}.

\begin{table}[h]
\centering
\scriptsize
\setlength{\tabcolsep}{2.8pt}
\renewcommand{\arraystretch}{0.92}
\caption{Full node-removal results. Acc@5/1 reports top-5 and top-1 accuracy;
Reg@5/1 reports regret at top-5 and top-1.}
\label{tab:app_node_remove_full}
\resizebox{\linewidth}{!}{%
\begin{tabular}{l l l l c c}
\toprule
Problem & Input & Loc. & Head & Acc@5/1 \(\uparrow\) & Reg@5/1 \(\downarrow\) \\
\midrule
ATSP100 & raw & N/A & MLP & 30.9\(\pm\)1.6 / 11.7\(\pm\)0.6 & 1.130\(\pm\)0.042 / 2.762\(\pm\)0.032 \\
ATSP100 & activation & L2 & Linear & 63.4\(\pm\)0.4 / 29.8\(\pm\)0.4 & 0.324\(\pm\)0.008 / 1.655\(\pm\)0.029 \\
ATSP100 & activation & Final & Linear & 58.9\(\pm\)0.4 / 26.0\(\pm\)0.6 & 0.415\(\pm\)0.010 / 1.740\(\pm\)0.023 \\
ATSP100 & activation & L2 & RowCol MLP & 69.8\(\pm\)0.4 / 36.1\(\pm\)1.0 & 0.229\(\pm\)0.005 / 1.363\(\pm\)0.039 \\
ATSP100 & activation & Final & RowCol MLP & 65.9\(\pm\)1.1 / 32.6\(\pm\)1.0 & 0.294\(\pm\)0.015 / 1.460\(\pm\)0.045 \\
ATSP100 & raw+act. & L2 & MLP & 69.6\(\pm\)0.7 / 34.9\(\pm\)0.7 & 0.228\(\pm\)0.010 / 1.376\(\pm\)0.038 \\
ATSP100 & raw+act. & Final & MLP & 66.7\(\pm\)1.1 / 34.1\(\pm\)1.3 & 0.277\(\pm\)0.018 / 1.433\(\pm\)0.045 \\
\midrule
ATSP500 & raw & N/A & MLP & 34.1\(\pm\)0.4 / 17.2\(\pm\)0.7 & 0.918\(\pm\)0.025 / 2.631\(\pm\)0.069 \\
ATSP500 & activation & L2 & Linear & 17.8\(\pm\)1.8 / 7.3\(\pm\)0.8 & 1.355\(\pm\)0.041 / 3.400\(\pm\)0.044 \\
ATSP500 & activation & Final & Linear & 16.6\(\pm\)0.6 / 7.6\(\pm\)0.3 & 1.358\(\pm\)0.030 / 3.290\(\pm\)0.042 \\
ATSP500 & activation & L2 & RowCol MLP & 23.4\(\pm\)1.0 / 11.3\(\pm\)0.4 & 1.112\(\pm\)0.025 / 3.036\(\pm\)0.042 \\
ATSP500 & activation & Final & RowCol MLP & 25.0\(\pm\)1.2 / 12.9\(\pm\)0.6 & 1.080\(\pm\)0.034 / 2.904\(\pm\)0.079 \\
ATSP500 & raw+act. & L2 & MLP & 33.4\(\pm\)0.9 / 16.6\(\pm\)0.6 & 0.907\(\pm\)0.020 / 2.735\(\pm\)0.051 \\
ATSP500 & raw+act. & Final & MLP & 32.9\(\pm\)1.0 / 16.7\(\pm\)0.9 & 0.917\(\pm\)0.022 / 2.703\(\pm\)0.062 \\
\midrule
RCVRP100 & raw & N/A & MLP & 8.0\(\pm\)0.4 / 1.8\(\pm\)0.1 & 6.064\(\pm\)0.053 / 11.884\(\pm\)0.073 \\
RCVRP100 & activation & L2 & Linear & 7.8\(\pm\)0.4 / 2.0\(\pm\)0.3 & 6.274\(\pm\)0.050 / 12.067\(\pm\)0.127 \\
RCVRP100 & activation & Final & Linear & 6.8\(\pm\)0.5 / 1.8\(\pm\)0.3 & 6.254\(\pm\)0.071 / 12.050\(\pm\)0.180 \\
RCVRP100 & activation & L2 & RowCol MLP & 7.6\(\pm\)0.4 / 1.9\(\pm\)0.3 & 6.261\(\pm\)0.113 / 12.150\(\pm\)0.200 \\
RCVRP100 & activation & Final & RowCol MLP & 7.1\(\pm\)0.5 / 2.0\(\pm\)0.2 & 6.225\(\pm\)0.064 / 12.087\(\pm\)0.162 \\
RCVRP100 & raw+act. & L2 & MLP & 8.1\(\pm\)0.4 / 2.1\(\pm\)0.2 & 6.064\(\pm\)0.053 / 11.737\(\pm\)0.079 \\
RCVRP100 & raw+act. & Final & MLP & 7.9\(\pm\)0.4 / 1.8\(\pm\)0.1 & 6.128\(\pm\)0.051 / 11.791\(\pm\)0.078 \\
\bottomrule
\end{tabular}%
}
\end{table}

\begin{table}[h]
\centering
\scriptsize
\setlength{\tabcolsep}{2.8pt}
\renewcommand{\arraystretch}{0.92}
\caption{Full edge-forbid results. RCVRP100 edge forbiddance is omitted from
the current scope because route-edge interventions in depot/customer multi-route
solutions require a separate task definition.}
\label{tab:app_edge_full}
\resizebox{\linewidth}{!}{%
\begin{tabular}{l l l l c c}
\toprule
Problem & Input & Loc. & Head & Acc@5/1 \(\uparrow\) & Reg@5/1 \(\downarrow\) \\
\midrule
ATSP100 & raw & N/A & MLP & 27.9\(\pm\)0.3 / 12.8\(\pm\)0.1 & 2.522\(\pm\)0.020 / 4.254\(\pm\)0.013 \\
ATSP100 & activation & L2 & Linear & 23.2\(\pm\)0.6 / 9.3\(\pm\)0.2 & 2.647\(\pm\)0.020 / 4.529\(\pm\)0.027 \\
ATSP100 & activation & Final & Linear & 16.5\(\pm\)0.5 / 5.6\(\pm\)0.3 & 2.959\(\pm\)0.040 / 5.014\(\pm\)0.035 \\
ATSP100 & activation & L2 & RowCol MLP & 25.8\(\pm\)0.5 / 12.3\(\pm\)0.6 & 2.545\(\pm\)0.041 / 4.238\(\pm\)0.043 \\
ATSP100 & activation & Final & RowCol MLP & 21.9\(\pm\)0.4 / 9.8\(\pm\)0.4 & 2.682\(\pm\)0.044 / 4.564\(\pm\)0.045 \\
ATSP100 & raw+act. & L2 & MLP & 29.7\(\pm\)0.9 / 15.9\(\pm\)0.3 & 2.366\(\pm\)0.057 / 3.865\(\pm\)0.039 \\
ATSP100 & raw+act. & Final & MLP & 29.3\(\pm\)0.6 / 15.6\(\pm\)0.4 & 2.399\(\pm\)0.052 / 3.878\(\pm\)0.035 \\
\midrule
ATSP500 & raw & N/A & MLP & 41.1\(\pm\)0.4 / 17.7\(\pm\)0.3 & 1.105\(\pm\)0.008 / 1.954\(\pm\)0.018 \\
ATSP500 & activation & L2 & Linear & 24.1\(\pm\)0.6 / 8.1\(\pm\)0.2 & 1.474\(\pm\)0.017 / 2.496\(\pm\)0.016 \\
ATSP500 & activation & Final & Linear & 16.6\(\pm\)0.3 / 6.9\(\pm\)0.3 & 1.681\(\pm\)0.025 / 2.700\(\pm\)0.034 \\
ATSP500 & activation & L2 & RowCol MLP & 37.8\(\pm\)0.9 / 16.3\(\pm\)0.5 & 1.155\(\pm\)0.033 / 1.975\(\pm\)0.036 \\
ATSP500 & activation & Final & RowCol MLP & 33.9\(\pm\)1.2 / 14.3\(\pm\)0.7 & 1.225\(\pm\)0.035 / 2.077\(\pm\)0.041 \\
ATSP500 & raw+act. & L2 & MLP & 45.9\(\pm\)0.6 / 22.4\(\pm\)0.7 & 1.055\(\pm\)0.037 / 1.693\(\pm\)0.036 \\
ATSP500 & raw+act. & Final & MLP & 46.0\(\pm\)0.8 / 21.5\(\pm\)0.6 & 1.028\(\pm\)0.022 / 1.716\(\pm\)0.040 \\
\bottomrule
\end{tabular}%
}
\end{table}

\begin{table}[h]
\centering
\scriptsize
\setlength{\tabcolsep}{3.0pt}
\renewcommand{\arraystretch}{0.95}
\caption{Full remove-5 results. The metric is average tour reduction after
removing five selected nodes and re-solving; higher is better.}
\label{tab:app_remove5_full}
\resizebox{0.85\linewidth}{!}{%
\begin{tabular}{l l l l c}
\toprule
Problem & Input & Loc. & Head & Tour reduction (\%) \(\uparrow\) \\
\midrule
ATSP100 & raw & N/A & MLP policy & 11.975\(\pm\)0.310 \\
ATSP100 & activation & L2 & Linear policy & 9.071\(\pm\)0.153 \\
ATSP100 & activation & Final & Linear policy & 8.306\(\pm\)0.387 \\
ATSP100 & activation & L2 & RowCol policy & 11.707\(\pm\)0.577 \\
ATSP100 & activation & Final & RowCol policy & 13.707\(\pm\)0.798 \\
ATSP100 & raw+act. & L2 & MLP policy & 13.316\(\pm\)0.182 \\
ATSP100 & raw+act. & Final & MLP policy & 14.659\(\pm\)0.285 \\
\midrule
ATSP500 & raw & N/A & MLP policy & 7.170\(\pm\)0.162 \\
ATSP500 & activation & L2 & Linear policy & 5.135\(\pm\)0.118 \\
ATSP500 & activation & Final & Linear policy & 5.294\(\pm\)0.263 \\
ATSP500 & activation & L2 & RowCol policy & 5.974\(\pm\)0.453 \\
ATSP500 & activation & Final & RowCol policy & 6.421\(\pm\)0.599 \\
ATSP500 & raw+act. & L2 & MLP policy & 7.318\(\pm\)0.129 \\
ATSP500 & raw+act. & Final & MLP policy & 7.253\(\pm\)0.113 \\
\bottomrule
\end{tabular}%
}
\end{table}

\newpage

\section{Heuristic Baselines}
\label{app:heuristic_results}

The main benchmark includes one deterministic local-repair heuristic per
problem-task pair. Table~\ref{tab:app_primary_heuristics} reports those heuristic rows, random references, and
additional heuristic variants. Accuracy values are percentages, and regret
values use the same normalized percentage-point label units as the main tables.

\begin{table}[h]
\centering
\scriptsize
\setlength{\tabcolsep}{3.0pt}
\renewcommand{\arraystretch}{0.95}
\caption{Primary local-repair heuristic baselines for node removal and edge
forbiddance.}
\label{tab:app_primary_heuristics}
\resizebox{\linewidth}{!}{%
\begin{tabular}{l l l c c c c}
\toprule
Task & Problem & Heuristic & \(n\) & Reg@5/1 \(\downarrow\) & Acc@5/1 \(\uparrow\) & Spearman \(\uparrow\) \\
\midrule
node-remove & ATSP100 & local\_delete\_saving & 1 & 0.408 / 1.121 & 65.5 / 42.4 & 0.364 \\
node-remove & ATSP500 & local\_delete\_saving & 1 & 1.180 / 2.877 & 19.5 / 9.1 & 0.068 \\
node-remove & RCVRP100 & route\_delete\_saving & 1 & 6.050 / 11.529 & 7.3 / 2.2 & 0.032 \\
edge-forbid & ATSP100 & edge\_replacement\_gap\_sum & 1 & 2.710 / 4.801 & 20.7 / 9.0 & 0.178 \\
edge-forbid & ATSP500 & edge\_replacement\_gap\_sum & 1 & 1.425 / 2.321 & 20.9 / 10.6 & 0.241 \\
\bottomrule
\end{tabular}%
}
\end{table}

\begin{table}[h]
\centering
\scriptsize
\setlength{\tabcolsep}{3.0pt}
\renewcommand{\arraystretch}{0.95}
\caption{Additional heuristic variants and random references.}
\label{tab:app_extra_heuristics}
\resizebox{\linewidth}{!}{%
\begin{tabular}{l l l c c c c}
\toprule
Task & Problem & Baseline & \(n\) & Reg@5/1 \(\downarrow\) & Acc@5/1 \(\uparrow\) & Spearman \(\uparrow\) \\
\midrule
edge-forbid & ATSP100 & edge\_cost & 1 & 4.236 / 6.110 & 3.9 / 0.7 & -0.052 \\
edge-forbid & ATSP100 & edge\_replacement\_gap\_min & 1 & 2.980 / 5.100 & 17.4 / 6.5 & 0.159 \\
edge-forbid & ATSP100 & random\_uniform & 10 & 3.965 / 5.908 & 5.0 / 1.1 & -0.001 \\
edge-forbid & ATSP500 & edge\_cost & 1 & 2.800 / 3.538 & 2.1 / 0.3 & -0.082 \\
edge-forbid & ATSP500 & edge\_replacement\_gap\_min & 1 & 1.760 / 2.767 & 13.0 / 4.3 & 0.217 \\
edge-forbid & ATSP500 & random\_uniform & 10 & 2.715 / 3.508 & 0.9 / 0.2 & -0.000 \\
node-remove & ATSP100 & random\_uniform & 10 & 2.527 / 4.017 & 5.1 / 1.0 & -0.000 \\
node-remove & ATSP500 & random\_uniform & 10 & 1.922 / 3.834 & 0.9 / 0.2 & 0.000 \\
node-remove & RCVRP100 & random\_uniform & 10 & 6.878 / 13.025 & 4.4 / 0.9 & -0.001 \\
\bottomrule
\end{tabular}%
}
\end{table}

\begin{table}[h]
\centering
\scriptsize
\setlength{\tabcolsep}{4.0pt}
\renewcommand{\arraystretch}{0.95}
\caption{Remove-5 local and random baselines.}
\label{tab:app_remove5_heuristics}
\begin{tabular}{l l c c c}
\toprule
Problem & Baseline & \(n\) & Instances & Tour improvement (\%) \(\uparrow\) \\
\midrule
ATSP100 & local\_delete\_saving\_top5 & 1 & 64 & 12.347 \\
ATSP100 & random\_top5 & 10 & 64 & 3.461 \(\pm\) 0.277 \\
ATSP500 & local\_delete\_saving\_top5 & 1 & 128 & 7.828 \\
ATSP500 & random\_top5 & 10 & 128 & 4.875 \(\pm\) 0.052 \\
\bottomrule
\end{tabular}
\end{table}

\section{Remove-5 RL Details}
\label{app:remove5_details}

Remove-5 is trained as an online intervention policy rather than a supervised
probe. At each episode, the policy observes candidate node features derived from
raw inputs, frozen activations, or their concatenation, and selects five nodes
without replacement. The reduced instance is then re-solved and the reward is
the resulting tour reduction percentage.

For raw-input policies, the policy head consumes manual features analogous to
the raw supervised probes. For activation policies, the policy consumes frozen
RRNCO encoder states. RowCol policies preserve separate row and column
activation streams before scoring candidates; MLP policies use flattened raw or
raw+activation features. Validation reward is used for model selection, and
final results are averaged over 10 seeds.

\section{Additional Probeability Analyses}
\label{app:probeability_details}

\paragraph{RL training dynamics.}
The training-dynamics figure in Section~\ref{sec:results_probeability} uses the
ATSP100 RRNCO foundation trajectory. We extract checkpoints throughout RL
continuation and train node-removal and edge-forbid probes at each checkpoint.
The plotted quantities are routing-model tour length, node-removal Acc@5 for
linear and row-column MLP probes, and edge-forbid Acc@5 for linear and
row-column MLP probes.

\paragraph{Iso-performance SL/RL sweep.}
The iso-performance sweep compares RRNCO checkpoints with approximately matched
route quality but different amounts of supervised pretraining before RL
continuation. The linear probe shows the clearest optimum at low but nonzero
supervised pretraining. Repeating the sweep with an MLP probe gives the same
qualitative Spearman pattern and similar, though less pronounced, trends in
Acc@5 and Reg@5. This supports the conclusion that the SL/RL effect is not only
a linear-readout artifact.

\section{Euclidean Cross-Architecture Comparison Details}
\label{app:cross_arch_details}

The Euclidean TSP100 comparison evaluates prescriptive probeability across
released checkpoints from representative NCO architecture families. This
comparison uses 100 Euclidean TSP100 instances with exact node-removal labels.
Probe metrics are averaged over three split/probe seeds, so Acc@1 is coarse and
should be interpreted directionally.

For constructive encoder-decoder models, we probe node-aligned encoder states
when available. For heatmap and diffusion models, the natural representation is
edge-centric, so we summarize edge hidden states into node-level inputs by
row/column pooling before training the probe. For LEHD, we probe the released
model's available node-level state; because LEHD shifts more computation into
the decoder, this may understate the information available in
partial-tour-conditioned decoder states. RRNCO-ATSP is included as an
out-of-domain point: it is trained and architecturally designed for real-road
asymmetric routing, not Euclidean TSP100.

\begin{table}[h]
\centering
\scriptsize
\setlength{\tabcolsep}{3.0pt}
\renewcommand{\arraystretch}{0.95}
\caption{Euclidean TSP100 cross-architecture comparison details. Policy gap is
reported against Concorde optima. Probe accuracies are mean \(\pm\) standard
deviation over three split/probe seeds.}
\label{tab:app_cross_arch}
\resizebox{\linewidth}{!}{%
\begin{tabular}{l l c c c p{0.38\linewidth}}
\toprule
Model & Family & Gap & Acc@1 & Acc@5 & Probed state \\
\midrule
DIFUSCO \citep{sun2023difusco}
& Diffusion & 0.24
& 50.0\(\pm\)10.0 & 93.3\(\pm\)11.5
& Final GNN denoiser edge hidden state, row/column mean pooled to nodes. \\
GCN-Heatmap \citep{joshi2019efficient}
& NAR GCN & 1.57
& 43.3\(\pm\)15.3 & 73.3\(\pm\)5.8
& Final residual-gated GCN edge state, row/column mean pooled to nodes. \\
POMO \citep{kwon2020pomo}
& AR constructive & 0.37
& 33.3\(\pm\)5.8 & 70.0\(\pm\)20.0
& TSP encoder output, per node. \\
Sym-NCO \citep{kim2022symnco}
& AR constructive & 0.38
& 36.7\(\pm\)5.8 & 83.3\(\pm\)5.8
& Sym-NCO-POMO encoder output, per node. \\
LEHD \citep{luo2023lehd}
& Heavy decoder & 0.56
& 16.7\(\pm\)15.3 & 40.0\(\pm\)10.0
& LEHD encoder output, per node. \\
RRNCO-ATSP \citep{son2025rrnco}
& AR & 3.82
& 50.0\(\pm\)0.0 & 83.3\(\pm\)5.8
& Policy encoder row and column embeddings concatenated per node. \\
AM \citep{kool2019attention}
& AR constructive & 4.57
& 20.0\(\pm\)0.0 & 60.0\(\pm\)10.0
& Graph attention encoder output, per node. \\
\bottomrule
\end{tabular}%
}
\end{table}


\newpage
\section*{NeurIPS Paper Checklist}

\begin{enumerate}

\item {\bf Claims}
    \item[] Question: Do the main claims made in the abstract and introduction accurately reflect the paper's contributions and scope?
    \item[] Answer: \answerYes{} 
    \item[] Justification: The abstract and introduction state the prescriptive probing contribution, real-road RRNCO scope, supervised/RL intervention tasks, and the qualified empirical claims.
    \item[] Guidelines:
    \begin{itemize}
        \item The answer \answerNA{} means that the abstract and introduction do not include the claims made in the paper.
        \item The abstract and/or introduction should clearly state the claims made, including the contributions made in the paper and important assumptions and limitations. A \answerNo{} or \answerNA{} answer to this question will not be perceived well by the reviewers. 
        \item The claims made should match theoretical and experimental results, and reflect how much the results can be expected to generalize to other settings. 
        \item It is fine to include aspirational goals as motivation as long as it is clear that these goals are not attained by the paper. 
    \end{itemize}

\item {\bf Limitations}
    \item[] Question: Does the paper discuss the limitations of the work performed by the authors?
    \item[] Answer: \answerYes{} 
    \item[] Justification: Section~\ref{sec:discussion} discusses solver-relative labels, counterfactual label cost, limited intervention families, and imperfect cross-architecture comparability.
    \item[] Guidelines:
    \begin{itemize}
        \item The answer \answerNA{} means that the paper has no limitation while the answer \answerNo{} means that the paper has limitations, but those are not discussed in the paper. 
        \item The authors are encouraged to create a separate ``Limitations'' section in their paper.
        \item The paper should point out any strong assumptions and how robust the results are to violations of these assumptions (e.g., independence assumptions, noiseless settings, model well-specification, asymptotic approximations only holding locally). The authors should reflect on how these assumptions might be violated in practice and what the implications would be.
        \item The authors should reflect on the scope of the claims made, e.g., if the approach was only tested on a few datasets or with a few runs. In general, empirical results often depend on implicit assumptions, which should be articulated.
        \item The authors should reflect on the factors that influence the performance of the approach. For example, a facial recognition algorithm may perform poorly when image resolution is low or images are taken in low lighting. Or a speech-to-text system might not be used reliably to provide closed captions for online lectures because it fails to handle technical jargon.
        \item The authors should discuss the computational efficiency of the proposed algorithms and how they scale with dataset size.
        \item If applicable, the authors should discuss possible limitations of their approach to address problems of privacy and fairness.
        \item While the authors might fear that complete honesty about limitations might be used by reviewers as grounds for rejection, a worse outcome might be that reviewers discover limitations that aren't acknowledged in the paper. The authors should use their best judgment and recognize that individual actions in favor of transparency play an important role in developing norms that preserve the integrity of the community. Reviewers will be specifically instructed to not penalize honesty concerning limitations.
    \end{itemize}

\item {\bf Theory assumptions and proofs}
    \item[] Question: For each theoretical result, does the paper provide the full set of assumptions and a complete (and correct) proof?
    \item[] Answer: \answerNA{} 
    \item[] Justification: The paper is empirical and does not introduce new theoretical results or proofs.
    \item[] Guidelines:
    \begin{itemize}
        \item The answer \answerNA{} means that the paper does not include theoretical results. 
        \item All the theorems, formulas, and proofs in the paper should be numbered and cross-referenced.
        \item All assumptions should be clearly stated or referenced in the statement of any theorems.
        \item The proofs can either appear in the main paper or the supplemental material, but if they appear in the supplemental material, the authors are encouraged to provide a short proof sketch to provide intuition. 
        \item Inversely, any informal proof provided in the core of the paper should be complemented by formal proofs provided in appendix or supplemental material.
        \item Theorems and Lemmas that the proof relies upon should be properly referenced. 
    \end{itemize}

    \item {\bf Experimental result reproducibility}
    \item[] Question: Does the paper fully disclose all the information needed to reproduce the main experimental results of the paper to the extent that it affects the main claims and/or conclusions of the paper (regardless of whether the code and data are provided or not)?
    \item[] Answer: \answerYes{}{} 
    \item[] Justification: Sections~\ref{sec:experiments} and Appendix~\ref{app:experiment_details} describe datasets, labels, splits, model/probe settings, training budgets, selection criteria, and evaluation metrics.
    \item[] Guidelines:
    \begin{itemize}
        \item The answer \answerNA{} means that the paper does not include experiments.
        \item If the paper includes experiments, a \answerNo{} answer to this question will not be perceived well by the reviewers: Making the paper reproducible is important, regardless of whether the code and data are provided or not.
        \item If the contribution is a dataset and\slash or model, the authors should describe the steps taken to make their results reproducible or verifiable. 
        \item Depending on the contribution, reproducibility can be accomplished in various ways. For example, if the contribution is a novel architecture, describing the architecture fully might suffice, or if the contribution is a specific model and empirical evaluation, it may be necessary to either make it possible for others to replicate the model with the same dataset, or provide access to the model. In general. releasing code and data is often one good way to accomplish this, but reproducibility can also be provided via detailed instructions for how to replicate the results, access to a hosted model (e.g., in the case of a large language model), releasing of a model checkpoint, or other means that are appropriate to the research performed.
        \item While NeurIPS does not require releasing code, the conference does require all submissions to provide some reasonable avenue for reproducibility, which may depend on the nature of the contribution. For example
        \begin{enumerate}
            \item If the contribution is primarily a new algorithm, the paper should make it clear how to reproduce that algorithm.
            \item If the contribution is primarily a new model architecture, the paper should describe the architecture clearly and fully.
            \item If the contribution is a new model (e.g., a large language model), then there should either be a way to access this model for reproducing the results or a way to reproduce the model (e.g., with an open-source dataset or instructions for how to construct the dataset).
            \item We recognize that reproducibility may be tricky in some cases, in which case authors are welcome to describe the particular way they provide for reproducibility. In the case of closed-source models, it may be that access to the model is limited in some way (e.g., to registered users), but it should be possible for other researchers to have some path to reproducing or verifying the results.
        \end{enumerate}
    \end{itemize}

\item {\bf Open access to data and code}
    \item[] Question: Does the paper provide open access to the data and code, with sufficient instructions to faithfully reproduce the main experimental results, as described in supplemental material?
    \item[] Answer: \answerYes{} 
    \item[] Justification: We provide an anonymized source-code supplement with
    setup instructions, experiment entrypoints, label-generation scripts,
    representation-extraction scripts, probe-training scripts, and table/figure
    summarization scripts. The archive is source-only and excludes generated
    datasets and checkpoints; these artifacts can be
    regenerated using the documented pipeline, and we plan to release the full
    public repository after review.
    \item[] Guidelines:
    \begin{itemize}
        \item The answer \answerNA{} means that paper does not include experiments requiring code.
        \item Please see the NeurIPS code and data submission guidelines (\url{https://neurips.cc/public/guides/CodeSubmissionPolicy}) for more details.
        \item While we encourage the release of code and data, we understand that this might not be possible, so \answerNo{} is an acceptable answer. Papers cannot be rejected simply for not including code, unless this is central to the contribution (e.g., for a new open-source benchmark).
        \item The instructions should contain the exact command and environment needed to run to reproduce the results. See the NeurIPS code and data submission guidelines (\url{https://neurips.cc/public/guides/CodeSubmissionPolicy}) for more details.
        \item The authors should provide instructions on data access and preparation, including how to access the raw data, preprocessed data, intermediate data, and generated data, etc.
        \item The authors should provide scripts to reproduce all experimental results for the new proposed method and baselines. If only a subset of experiments are reproducible, they should state which ones are omitted from the script and why.
        \item At submission time, to preserve anonymity, the authors should release anonymized versions (if applicable).
        \item Providing as much information as possible in supplemental material (appended to the paper) is recommended, but including URLs to data and code is permitted.
    \end{itemize}

\item {\bf Experimental setting/details}
    \item[] Question: Does the paper specify all the training and test details (e.g., data splits, hyperparameters, how they were chosen, type of optimizer) necessary to understand the results?
    \item[] Answer: \answerYes{} 
    \item[] Justification: The paper specifies data sources, splits, label generation, solvers, probe locations, probe heads, hyperparameter selection, optimizer settings, and evaluation metrics in the main text and appendix.
    \item[] Guidelines:
    \begin{itemize}
        \item The answer \answerNA{} means that the paper does not include experiments.
        \item The experimental setting should be presented in the core of the paper to a level of detail that is necessary to appreciate the results and make sense of them.
        \item The full details can be provided either with the code, in appendix, or as supplemental material.
    \end{itemize}

\item {\bf Experiment statistical significance}
    \item[] Question: Does the paper report error bars suitably and correctly defined or other appropriate information about the statistical significance of the experiments?
    \item[] Answer: \answerYes{} 
    \item[] Justification: Main learned results are averaged over 10 seeds, with full mean \(\pm\) standard deviation tables in Appendix~\ref{app:full_main_results}; cross-architecture results report mean \(\pm\) standard deviation over three split/probe seeds.
    \item[] Guidelines:
    \begin{itemize}
        \item The answer \answerNA{} means that the paper does not include experiments.
        \item The authors should answer \answerYes{} if the results are accompanied by error bars, confidence intervals, or statistical significance tests, at least for the experiments that support the main claims of the paper.
        \item The factors of variability that the error bars are capturing should be clearly stated (for example, train/test split, initialization, random drawing of some parameter, or overall run with given experimental conditions).
        \item The method for calculating the error bars should be explained (closed form formula, call to a library function, bootstrap, etc.)
        \item The assumptions made should be given (e.g., Normally distributed errors).
        \item It should be clear whether the error bar is the standard deviation or the standard error of the mean.
        \item It is OK to report 1-sigma error bars, but one should state it. The authors should preferably report a 2-sigma error bar than state that they have a 96\% CI, if the hypothesis of Normality of errors is not verified.
        \item For asymmetric distributions, the authors should be careful not to show in tables or figures symmetric error bars that would yield results that are out of range (e.g., negative error rates).
        \item If error bars are reported in tables or plots, the authors should explain in the text how they were calculated and reference the corresponding figures or tables in the text.
    \end{itemize}

\item {\bf Experiments compute resources}
    \item[] Question: For each experiment, does the paper provide sufficient information on the computer resources (type of compute workers, memory, time of execution) needed to reproduce the experiments?
    \item[] Answer: \answerYes{} 
    \item[] Justification: Appendix~\ref{app:experiment_details} reports the H200 compute setting and wall-clock budgets for foundation training, SL/RL sweeps, remove-5 RL heads, and probe training.
    \item[] Guidelines:
    \begin{itemize}
        \item The answer \answerNA{} means that the paper does not include experiments.
        \item The paper should indicate the type of compute workers CPU or GPU, internal cluster, or cloud provider, including relevant memory and storage.
        \item The paper should provide the amount of compute required for each of the individual experimental runs as well as estimate the total compute. 
        \item The paper should disclose whether the full research project required more compute than the experiments reported in the paper (e.g., preliminary or failed experiments that didn't make it into the paper). 
    \end{itemize}
    
\item {\bf Code of ethics}
    \item[] Question: Does the research conducted in the paper conform, in every respect, with the NeurIPS Code of Ethics \url{https://neurips.cc/public/EthicsGuidelines}?
    \item[] Answer: \answerYes{} 
    \item[] Justification: The work uses road network data; it does not involve human subjects, sensitive personal data, or high-risk deployment.
    \item[] Guidelines:
    \begin{itemize}
        \item The answer \answerNA{} means that the authors have not reviewed the NeurIPS Code of Ethics.
        \item If the authors answer \answerNo, they should explain the special circumstances that require a deviation from the Code of Ethics.
        \item The authors should make sure to preserve anonymity (e.g., if there is a special consideration due to laws or regulations in their jurisdiction).
    \end{itemize}

\item {\bf Broader impacts}
    \item[] Question: Does the paper discuss both potential positive societal impacts and negative societal impacts of the work performed?
    \item[] Answer: \answerYes{} 
    \item[] Justification: The paper discusses operational decision-support use cases and limitations. Positive impacts include more efficient what-if analysis; possible negative impacts include overreliance on solver-relative predictions if used without validation.
    \item[] Guidelines:
    \begin{itemize}
        \item The answer \answerNA{} means that there is no societal impact of the work performed.
        \item If the authors answer \answerNA{} or \answerNo, they should explain why their work has no societal impact or why the paper does not address societal impact.
        \item Examples of negative societal impacts include potential malicious or unintended uses (e.g., disinformation, generating fake profiles, surveillance), fairness considerations (e.g., deployment of technologies that could make decisions that unfairly impact specific groups), privacy considerations, and security considerations.
        \item The conference expects that many papers will be foundational research and not tied to particular applications, let alone deployments. However, if there is a direct path to any negative applications, the authors should point it out. For example, it is legitimate to point out that an improvement in the quality of generative models could be used to generate Deepfakes for disinformation. On the other hand, it is not needed to point out that a generic algorithm for optimizing neural networks could enable people to train models that generate Deepfakes faster.
        \item The authors should consider possible harms that could arise when the technology is being used as intended and functioning correctly, harms that could arise when the technology is being used as intended but gives incorrect results, and harms following from (intentional or unintentional) misuse of the technology.
        \item If there are negative societal impacts, the authors could also discuss possible mitigation strategies (e.g., gated release of models, providing defenses in addition to attacks, mechanisms for monitoring misuse, mechanisms to monitor how a system learns from feedback over time, improving the efficiency and accessibility of ML).
    \end{itemize}
    
\item {\bf Safeguards}
    \item[] Question: Does the paper describe safeguards that have been put in place for responsible release of data or models that have a high risk for misuse (e.g., pre-trained language models, image generators, or scraped datasets)?
    \item[] Answer: \answerYes{} 
    \item[] Justification: The paper does not release high-risk generative models, scraped datasets, or dual-use systems requiring special safeguards.
    \item[] Guidelines:
    \begin{itemize}
        \item The answer \answerNA{} means that the paper poses no such risks.
        \item Released models that have a high risk for misuse or dual-use should be released with necessary safeguards to allow for controlled use of the model, for example by requiring that users adhere to usage guidelines or restrictions to access the model or implementing safety filters. 
        \item Datasets that have been scraped from the Internet could pose safety risks. The authors should describe how they avoided releasing unsafe images.
        \item We recognize that providing effective safeguards is challenging, and many papers do not require this, but we encourage authors to take this into account and make a best faith effort.
    \end{itemize}

\item {\bf Licenses for existing assets}
    \item[] Question: Are the creators or original owners of assets (e.g., code, data, models), used in the paper, properly credited and are the license and terms of use explicitly mentioned and properly respected?
    \item[] Answer: \answerYes{} 
    \item[] Justification: The paper cites the existing models, solvers, datasets, and released checkpoints used in the experiments, including RRNCO, LKH/LKH3, and external NCO checkpoints.
    \item[] Guidelines:
    \begin{itemize}
        \item The answer \answerNA{} means that the paper does not use existing assets.
        \item The authors should cite the original paper that produced the code package or dataset.
        \item The authors should state which version of the asset is used and, if possible, include a URL.
        \item The name of the license (e.g., CC-BY 4.0) should be included for each asset.
        \item For scraped data from a particular source (e.g., website), the copyright and terms of service of that source should be provided.
        \item If assets are released, the license, copyright information, and terms of use in the package should be provided. For popular datasets, \url{paperswithcode.com/datasets} has curated licenses for some datasets. Their licensing guide can help determine the license of a dataset.
        \item For existing datasets that are re-packaged, both the original license and the license of the derived asset (if it has changed) should be provided.
        \item If this information is not available online, the authors are encouraged to reach out to the asset's creators.
    \end{itemize}

\item {\bf New assets}
    \item[] Question: Are new assets introduced in the paper well documented and is the documentation provided alongside the assets?
    \item[] Answer: \answerNA{} 
    \item[] Justification: The submission does not currently release a new dataset, benchmark suite, or model as a paper artifact.
    \item[] Guidelines:
    \begin{itemize}
        \item The answer \answerNA{} means that the paper does not release new assets.
        \item Researchers should communicate the details of the dataset\slash code\slash model as part of their submissions via structured templates. This includes details about training, license, limitations, etc. 
        \item The paper should discuss whether and how consent was obtained from people whose asset is used.
        \item At submission time, remember to anonymize your assets (if applicable). You can either create an anonymized URL or include an anonymized zip file.
    \end{itemize}

\item {\bf Crowdsourcing and research with human subjects}
    \item[] Question: For crowdsourcing experiments and research with human subjects, does the paper include the full text of instructions given to participants and screenshots, if applicable, as well as details about compensation (if any)? 
    \item[] Answer: \answerNA{} 
    \item[] Justification: The work does not involve crowdsourcing or human-subject experiments.
    \item[] Guidelines:
    \begin{itemize}
        \item The answer \answerNA{} means that the paper does not involve crowdsourcing nor research with human subjects.
        \item Including this information in the supplemental material is fine, but if the main contribution of the paper involves human subjects, then as much detail as possible should be included in the main paper. 
        \item According to the NeurIPS Code of Ethics, workers involved in data collection, curation, or other labor should be paid at least the minimum wage in the country of the data collector. 
    \end{itemize}

\item {\bf Institutional review board (IRB) approvals or equivalent for research with human subjects}
    \item[] Question: Does the paper describe potential risks incurred by study participants, whether such risks were disclosed to the subjects, and whether Institutional Review Board (IRB) approvals (or an equivalent approval/review based on the requirements of your country or institution) were obtained?
    \item[] Answer: \answerNA{} 
    \item[] Justification: The work does not involve human-subject research.
    \item[] Guidelines:
    \begin{itemize}
        \item The answer \answerNA{} means that the paper does not involve crowdsourcing nor research with human subjects.
        \item Depending on the country in which research is conducted, IRB approval (or equivalent) may be required for any human subjects research. If you obtained IRB approval, you should clearly state this in the paper. 
        \item We recognize that the procedures for this may vary significantly between institutions and locations, and we expect authors to adhere to the NeurIPS Code of Ethics and the guidelines for their institution. 
        \item For initial submissions, do not include any information that would break anonymity (if applicable), such as the institution conducting the review.
    \end{itemize}

\item {\bf Declaration of LLM usage}
    \item[] Question: Does the paper describe the usage of LLMs if it is an important, original, or non-standard component of the core methods in this research? Note that if the LLM is used only for writing, editing, or formatting purposes and does \emph{not} impact the core methodology, scientific rigor, or originality of the research, declaration is not required.
    \item[] Answer: \answerNA{} 
    \item[] Justification: LLMs are not used as a core methodological component of the research.    \item[] Guidelines:
    \begin{itemize}
        \item The answer \answerNA{} means that the core method development in this research does not involve LLMs as any important, original, or non-standard components.
        \item Please refer to our LLM policy in the NeurIPS handbook for what should or should not be described.
    \end{itemize}

\end{enumerate}

\end{document}